\pdfoutput=1

\documentclass[11pt]{article}

\usepackage[preprint]{acl}

\usepackage{times}
\usepackage{latexsym}

\usepackage{microtype}

\usepackage{inconsolata}

\usepackage{graphicx}

\usepackage[utf8]{inputenc} 
\usepackage[T1]{fontenc}    
\usepackage{hyperref}       
\usepackage{url}            
\usepackage{booktabs}       
\usepackage{amsfonts}       
\usepackage{nicefrac}       
\usepackage{microtype}      
\usepackage{xcolor}         

\usepackage{graphicx}
\usepackage{subfigure}

\usepackage{amsmath}
\usepackage{bm}
\usepackage{graphicx} 
\usepackage{wrapfig} 
\usepackage{bbold}

\usepackage{tcolorbox}
\usepackage{multirow}
\usepackage{adjustbox}

\title{Graph Neural Network Enhanced Retrieval for Question Answering of Large Language Models}

\author{%
Zijian Li$^{12}$\thanks{Work done during an internship at Microsoft Research Asia.
}~~,
Qingyan Guo$^{23}$, Jiawei Shao$^{1}$, Lei Song$^{2}$, Jiang Bian$^{2}$, Jun Zhang$^{1}$\thanks{Corresponding Author.}~~, Rui Wang$^{2\dag}$\\
$^1$Hong Kong University of Science and Technology\quad 
$^2$Microsoft Research\quad 
$^3$Tsinghua University\\
\texttt{\{zijian.li,jiawei.shao\}@connect.ust.hk}, 
\texttt{gqy22@mails.tsinghua.edu.cn},\\
\texttt{\{lesong,jiabia,ruiwa\}@microsoft.com}, 
\texttt{eejzhang@ust.hk}
}

\begin{document}
\maketitle
\begin{abstract}
Retrieval augmented generation has revolutionized large language model (LLM) outputs by providing factual supports.
Nevertheless, it struggles to capture all the necessary knowledge for complex reasoning questions.
  Existing retrieval methods typically divide reference documents into passages, treating them in isolation. These passages, however, are often interrelated, such as passages that are contiguous or share the same keywords. Therefore, it is crucial to recognize such relatedness for enhancing the retrieval process. 
  In this paper, we propose a novel retrieval method, called \textit{GNN-Ret}, which leverages \textit{graph neural networks} (GNNs) to enhance retrieval by exploiting the relatedness between passages.
  Specifically, we first construct a \textit{graph of passages} by connecting passages that are structure-related or keyword-related.
  A \textit{graph neural network} (GNN) is then leveraged to exploit the relationships between passages and improve the retrieval of supporting passages.
  Furthermore, we extend our method to handle multi-hop reasoning questions using a \textit{recurrent graph neural network} (RGNN), named \textit{RGNN-Ret}. 
  At each step, \textit{RGNN-Ret} integrates the graphs of passages from previous steps, thereby enhancing the retrieval of supporting passages. 
  Extensive experiments on benchmark datasets demonstrate that \textit{GNN-Ret} achieves higher accuracy for question answering with a single query of LLMs than strong baselines that require multiple queries, and \textit{RGNN-Ret} further improves accuracy and achieves state-of-the-art performance, with up to 10.4$\%$ accuracy improvement on the 2WikiMQA dataset.
\end{abstract}

\section{Introduction}
\label{intro}

Large language models (LLMs) continue to struggle with factual errors when encountering knowledge intensive questions \citep{huang2023survey, mallen2022not,ji2023survey}.
Although retrieval-augmented LLMs \citep{rag} have improved factuality and precision of question answering by including relevant passages, there remains a persistent challenge in accurately capturing all the supporting passages when encountering complex knowledge-intensive questions.
This limitation can be attributed to the inherent information asymmetry in complex questions. In particular, the questions tend to consist of elaborated background details, leaving only a small portion dedicated to specific inquiries.
As an example, in question: \textit{`Why did crime rise on Mars after the Mafia's arrival?'}, the majority part delves into the background (i.e., \textit{`crime rise ...... Mafia's arrival'}) with only a few words requesting the reason (i.e., \textit{`Why'}), which consequently retrieves passages on details of crime rising instead of the reason for it.

\begin{figure*}
    \centering
    \vspace{-0.3in}
    \includegraphics[width=0.95\textwidth]{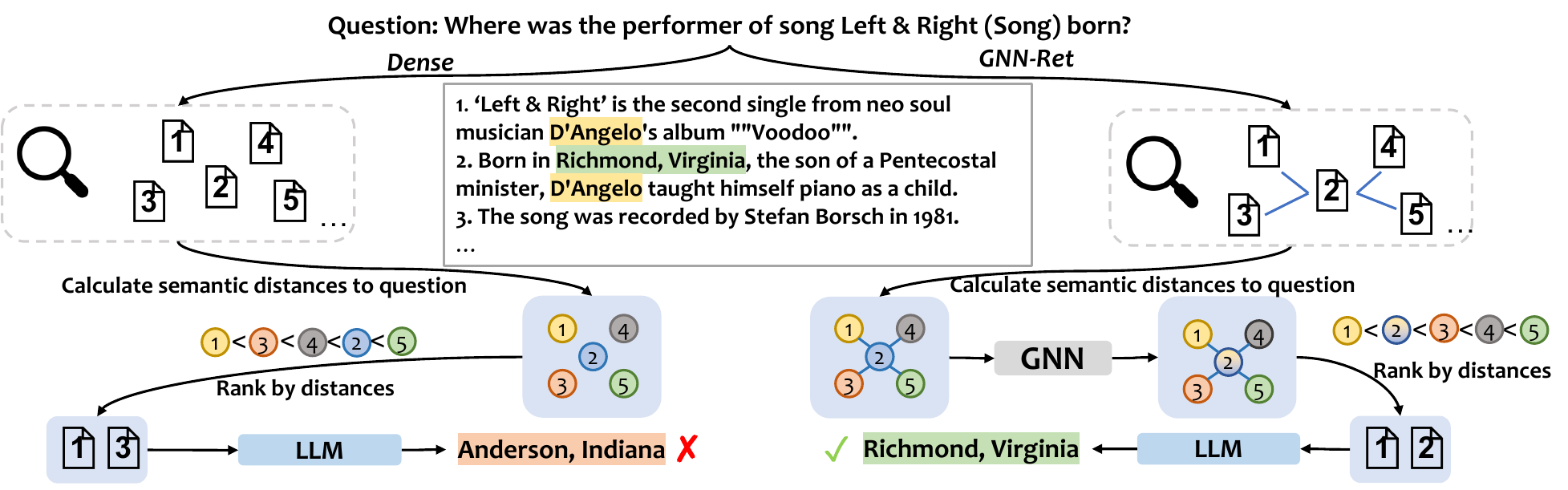}
    \vspace{-0.15in}
    \caption{Overview of comparison between dense retrieval and \textit{GNN-Ret}. The shared keywords and ground-truth answers are highlighted in {\color{yellow} yellow} and {\color{green} green}, respectively. By considering the relatedness between passages, \textit{GNN-Ret} can retrieve all the supporting passages for QA.}
    \label{fig:overview}
\vspace{-0.2in}
\end{figure*}

This phenomenon also frequently arises in multi-hop reasoning questions.
Considering the sample question `\textit{Where was the performer of song Left \& Right (Song) born?}' in Fig. \ref{fig:overview}, it becomes apparent that while we can retrieve the knowledge that \textit{the performer of song Left \& Right is} \textit{D'Angelo}, his birthplace remains absent.
Previous works \cite{ircot, self_ask, yao2022react} have attempted to address this issue by incorporating multi-hop reasoning or question rewriting into retrieval processes, which enables them to retrieve information based on prior reasoning outcomes.
LLMs, however, often generate plausible reasons and incorrect rewriting questions without accurate prior domain knowledge of the given question \cite{huang2023survey, zhang2023siren}, thus affecting the subsequent retrieval process.

One reason why existing methods struggle to handle the information asymmetry is their tendency to consider passages in isolation \cite{karpukhin2020dense} and retrieve the supporting passages based mainly on semantic distances, making it difficult to retrieve all the supporting passages, especially those containing only few words for inquiry.
However, the supporting passages of background and inquiry information are usually correlated.
They can be \textit{structure-related} when located in the same section or document, e.g., the happened event (i.e., \textit{crime rises on Mars}) and its corresponding reason are located in the same section but different passages.
Also, they can be \textit{keyword-related} by sharing the same keywords or entities, e.g., the passages of background information (\textit{performer of song Left $\&$ Right}) and inquiry information (\textit{birthplace of D'Angelo}) in Fig. \ref{fig:overview} share the same keyword: \textit{D'Angelo}.
By considering the relatedness between passages, it is possible to retrieve all the supporting passages even when they have significant semantic differences from the query.
In this work, we aim to enhance retrieval by taking the relatedness between passages into account.

\textbf{Contributions.}
To establish the relatedness between passages for retrieval purposes, this study initially constructs a \textit{graph of passages} by connecting individual passages based on both \textit{structural information} and \textit{shared keywords}, with each passage as a node in this graph. 
The key challenge lies in how to effectively leverage the passage of graphs to enhance retrieval coverage.
Graph neural networks (GNNs) are neural networks tailored to analyze graph data and adeptly grasp the relationships between nodes and edges \cite{gnn}. 
Thus, we propose to leverage a GNN to enhance the retrieval process by effectively capturing the relatedness between passages and name this method \textit{GNN-Ret}. 
The GNN facilitates the integration of semantic distances between related passages, thus enabling the semantic distances of passages containing background information to impact the retrieval of passages relevant to the inquiry. 
To address multi-hop reasoning questions, we further propose a retrieval method, named \textit{RGNN-Ret}, which leverages a \textit{recurrent graph neural network} (RGNN) to enhance
the retrieval process at each step by integrating the retrievals from previous steps through the interconnected graphs of passages. 
This boosts the retrieval coverage for supporting passages over steps even when LLMs generate incorrect reasons or subquestions for retrieval.

\begin{figure}[h]
\vspace{-0.1in}
    \centering
    \includegraphics[width=0.9\linewidth]{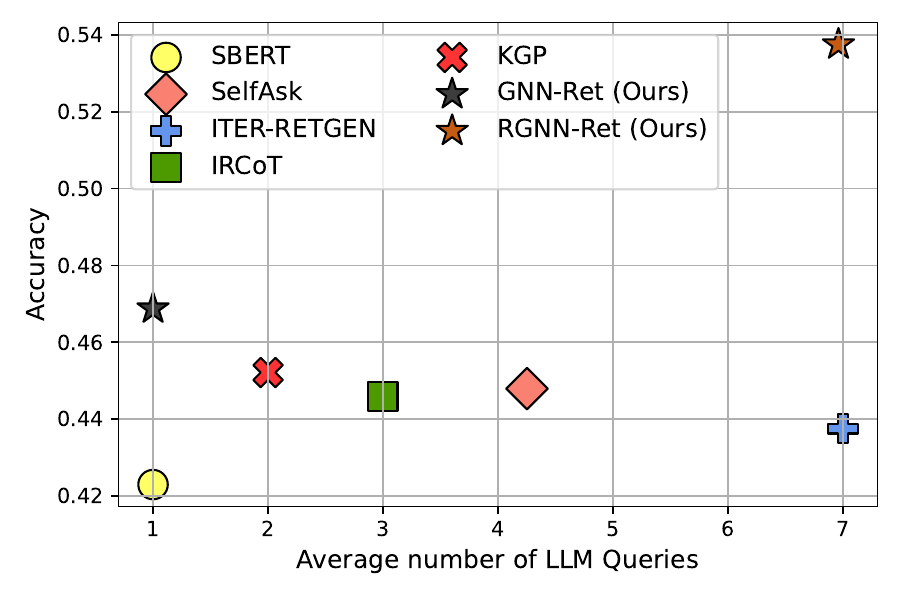}
    \vspace{-0.2in}
    \caption{Accuracy and average number of LLM queries for our proposed methods and baselines on 2WikiMQA.}
    \label{trade_off}
\end{figure}

Through the experiments on four benchmark datasets, we demonstrate the effectiveness of GNNs in enhancing retrieval of supporting passages and thus improving accuracy for QA.
For example, as shown in Fig. \ref{trade_off}, our proposed \textit{GNN-Ret} with a single query of LLMs significantly outperforms baselines in accuracy on 2WikiMQA \cite{2wiki}, including those methods that require multiple queries of LLMs.
Moreover, by extending GNN to multi-hop reasoning questions, our proposed \textit{RGNN-Ret} achieves state-of-the-art accuracy.

\section{Related Works}
\textbf{Retrieval-augmented LLM.}
QA \cite{voorhees1999trec} is a task that often requires external and up-to-date knowledge sources to answer factoid-based questions.
Retrieve-and-read is the basic framework for these questions \cite{gao2023retrieval, zhu2023large}.
For retrieval, sparse retrievers, e.g., TF-IDF and BM25 \cite{bm25}, or dense retrievers, e.g., DPR \cite{dpr} and Contriever \cite{contriver}, are applied to compute lexical distances or semantic distances between passages and the question.
The passages with smaller distances are retrieved, which are then prefixed with the question for factual answering \cite{rag}.
However, this framework treats passages as isolated units, making it difficult to retrieve all the supporting passages for the question in a single step.
To address this limitation, many approaches have been proposed to integrate the retrieval and reasoning processes to improve the retrieval of supporting passages \cite{gao2023retrieval,iter_retgen, ITRG, ircot, flare, yao2022react, self_ask}.
They first prompt LLMs to generate the next-step reason or subquestion and then use it to guide retrieval for QA.
However, a challenge in these methodologies pertains to the potential generation of hallucinated reasons or erroneous subquestions by LLMs without accurate domain knowledge of this question, leading to degraded subsequent retrievals \cite{huang2023survey}.

\textbf{Graph-enhanced LLM.}
As structured, explicit, and editable representations of knowledge, knowledge graphs (KGs) have been effectively used for boosting retrieval coverage \cite{min2019knowledge} and enhancing reasoning capabilities of LLMs \cite{li2023chain, xie2022unifiedskg, baek2023knowledge, wang2023boosting, park2023graph, ding2024enhancing, jin2023large, li2023survey, xiong2024teilp,he2024g}.
Instead of naively combining KGs with LLMs, many works directly prompt LLMs to perform reasoning on KGs \cite{sun2023think, StructGPT, luo2023reasoning}.
They first identified the entities on the knowledge graph and then search reasoning paths accordingly \cite{sun2023think, luo2023reasoning, StructGPT}.
However, KG-based methods encounter a limitation that they can only handle questions that are effectively represented as KGs. They may struggle with questions that necessitate a comprehensive understanding of lengthy contextual information, such as the mentioned example: the cause of crime increasing.
Rather than KGs, a recent study \cite{jin2024graph} attempts to construct a graph of passages and perform reasoning on this graph.
It connects documents according to some specific words and performs reasoning on this graph for QA. 
In contrast, we utilize the graph of passages to enhance the retrieval process.
Another related work, KGP \cite{wang2024knowledge}, performs retrieval on the graph and search related passages.
However, this approach does not consider the integration between related passages and may not effectively enhance retrieval for the supporting passages with poor semantic similarity.

\section{Graph Neural Network Enhanced Retrieval for QA}

The QA system typically adopts a \textit{retrieve-and-read} routine \cite{rag}: the retriever computes semantic distances between passages and the question in the embedding space, finds out relevant passages from the whole corpus, and then the \textit{reader} (LLM) generates the answer based on them.
Considering the information asymmetry of complex questions, directly using the initial semantic distances of passages for retrieval is difficult to retrieve all the supporting passages.
Therefore, we propose to process by taking the relatedness between passages into account.
Specifically, we construct a \textit{graph of passages} (GoPs) by connecting related passages (Section \ref{graph construction}).
To exploit the relatedness between passages to enhance retrieval, we utilize a GNN to enable integration between related passages (Section \ref{GNN}).
For multi-hop questions, we further leverage an RGNN to establish relationship between graphs over steps to enhance retrieval over answering processes (Section \ref{RGNN}).

\subsection{Graph of Passages (GoPs)}
\label{graph construction}

\textbf{Establish relationships between related passages.}
Existing retrieval methods rely on semantic distances between passages and the question \cite{gao2023retrieval, zhu2023large}, which may fail to retrieve some supporting passages with large semantic distances, but related to those with small semantic distances to the question, due to the information asymmetry.
These passages can be \textit{structural-related}, e.g., the happened event (\textit{`crime rising'}) and its corresponding reasons are located in the same section.
As shown in Fig. \ref{fig:overview}, they can also be \textit{keyword-related}, e.g., the supporting passages for background information (`\textit{performer of song Left $\&$ Right}') and inquiry information (`\textit{the birthplace of D'Angelo}') share the same keyword (`\textit{D'Angelo}').
Therefore, we propose to establish relationship between passages using structural information and shared keywords to construct the graph of passages.
Specifically, when chunking the documents, we record the order of passages and connect passages that are physically next to each other in documents.
In addition, we extract keywords from passages by prompting LLMs and connect passages containing the same keywords \cite{min2019knowledge, wang2024knowledge}.
As such, the related passages are connected, having the potential to facilitate retrieval.

\subsection{GNN Enhanced Retrieval on GoPs}
\label{GNN}
After establishing connections between related passages, we obtain a GoPs, with each passage as a node.
When conducting the retrieval process, we first compute the semantic distances between passages and the question $q$ \cite{gao2023retrieval, zhu2023large}.
The semantic distances of related passages are connected according to the GoPs.
To take relatedness between passages into account, we utilize a GNN to process semantic distances based on GoPs and obtain the \textit{integrated semantic distances}, which benefits the retrieval of supporting passages.

\textbf{Graph neural network.}
A graph is defined as an ordered pair $\mathcal{G}=\{\mathcal{V}, \mathcal{E}\}$, where  $\mathcal{V}$ is the set of nodes $v_i$ and $\mathcal{E}$ is the set of edges $e_{ij}$ connecting two nodes $v_i$ and $v_j$.
We define $h_i^l$ as the hidden state (integrated semantic distance) of the $i-$th node for $i \in \{1,\cdots,|\mathcal{V}|\}$ at layer $l$.
We input semantic distances of passages into GNN as $h_i^0$.
The set of nodes $\mathcal{N}_{v_i}$ stands for the neighbors of node $v_i$. 
For an $L-$layer GNN \cite{gnn}, the hidden state of each node $v_i \in \mathcal{V}$ is updated through iterative interactions with neighbors.
Specifically, at layer $l$, the received messages $m_i^l$ of node $v_i$ are calculated using the hidden states of its neighbors. Then the received messages $m_i^l$ and the previous-layer hidden state $h_i^{l-1}$ are utilized to compute the hidden state $h_i^{l}$ at layer $l$.
By processing hidden states for all the passages layer by layer, we obtain the integrated semantic distance $h_i^L$ for each node $v_i \in \mathcal{V}$.
We elaborately introduce GNNs for retrieval as below.

\textbf{Minimum semantic distance as message.}
Since each passage maintains and shares many keywords with other passages, each passage has a large number of neighbors.
Some of them are not relevant to the question, contributing to large semantic distances and misleading information.
Therefore, each node $v_i$ only receives the message from the neighbor that has the minimum (integrated) semantic distance to the question at layer $l$.
The received message of node $v_i$ is thus formulated as $ m_i^l = \min_{j \in \mathcal{N}_{v_i}} h_j^{l-1}$.

\begin{figure*}
\vspace{-0.3in}
    \centering
    \includegraphics[width=0.9\textwidth]{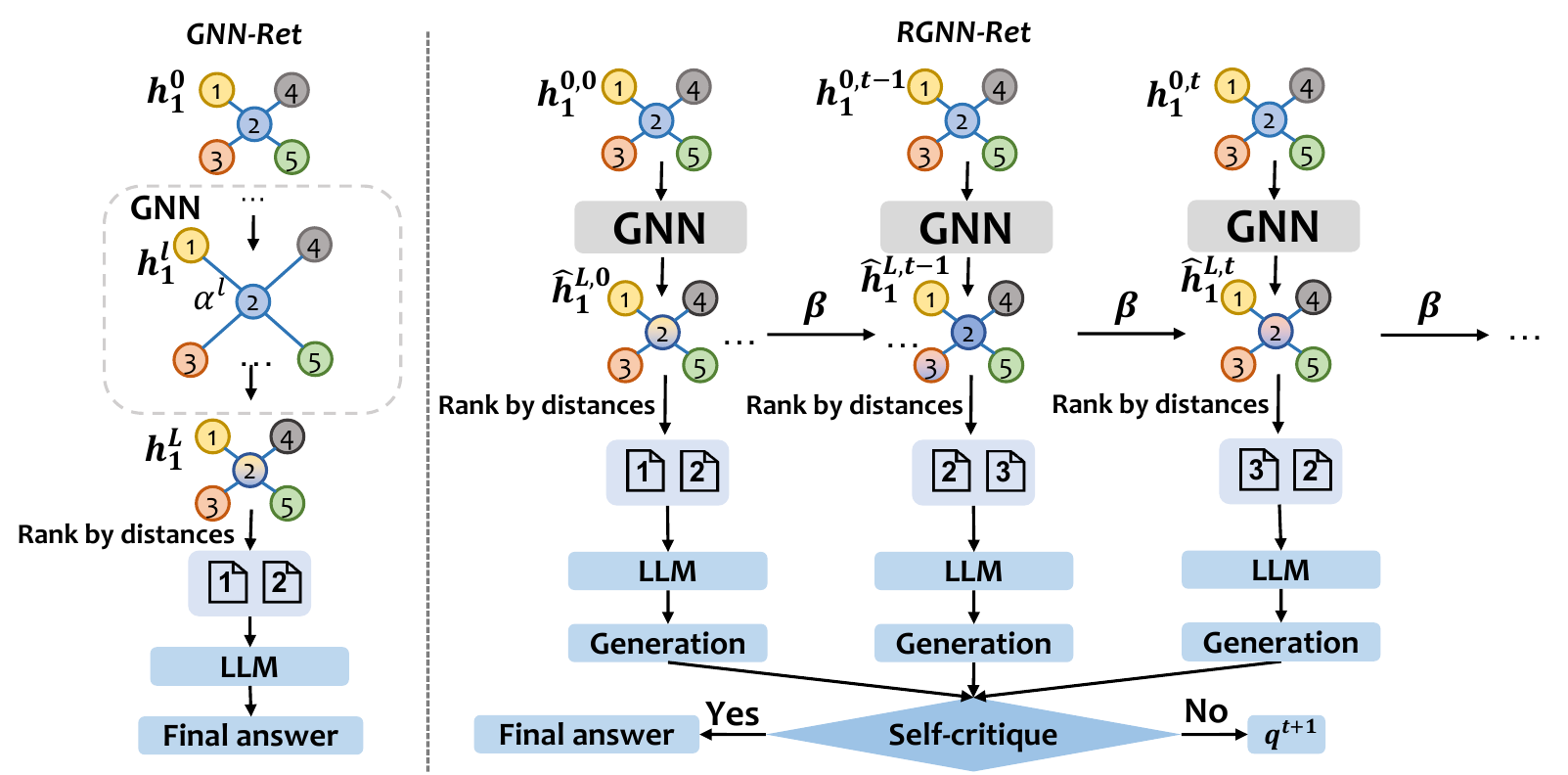}
        \vspace{-0.1in}
    \caption{Illustration of \textit{GNN-Ret} and \textit{RGNN-Ret}.}
    \label{fig:illustration}
        \vspace{-0.2in}
\end{figure*}
    
\textbf{Relevant nodes sampling.}
Given the large scale of GoPs with plenty of nodes, it is computationally inefficient to propagate information across all of them.
Besides, allowing irrelevant nodes to pass messages to their neighbors may affect the retrieval process.
Therefore, we only sample the relevant node set $\mathcal{T}_K^{l}$ with the top $K$ smallest (integrated) semantic distances at layer $l$.
Only neighbors of relevant nodes receive messages from them and update hidden states.
We define $\mathcal{S}_K^l$ as the neighbor set of the relevant node set $\mathcal{T}_K^l$.
For each node $v_i \in \mathcal{S}_K^{l}$, the message passed from neighbors that are relevant nodes is reformulated as follows:
\begin{equation}
\label{min}
\setlength\abovedisplayskip{3pt}
\setlength\belowdisplayskip{3pt}
    m_i^l = \min_{j \in \mathcal{N}_{v_i}, v_j\in \mathcal{T}_K^{l}} h_j^{l-1}.
\end{equation}
By integrating the message from neighbors, the hidden state of each node $v_i \in \mathcal{S}_K^l$ at layer $l$ is computed using the parameter $\alpha^l$ as follows: 
\begin{equation}
\label{integration}
\setlength\abovedisplayskip{3pt}
\setlength\belowdisplayskip{3pt}
    h_i^l = \alpha^l h_i^{l-1} + (1-\alpha^l) m_i^{l}.
\end{equation}
The nodes without received messages maintain their hidden states of the previous layer.
For the GNN with $L$ layers, we compute the hidden state layer by layer and then obtain the integrated semantic distance $h_i^{L}$ for each node $v_i$. 
With GNN, the supporting passages for inquiry information integrate with the small semantic distance from those for background information and thereby obtain smaller integrated semantic distances, promoting them to be retrieved.

\textbf{Hinge objective for GNN.}
To retrieve all the supporting passages, we aim to reduce their integrated semantic distances more than those of non-target nodes.
With the ground-truth set of supporting passages $\mathcal{S}_Y$, we first define the average integrated semantic distances of supporting passages and non-target nodes as $\bar{d}_{Y}^L = \frac{1}{|\mathcal{S}_{Y}|} \sum_{j \in \mathcal{S}_Y} h_j^L$ and $\bar{d}_{O/Y}^L = \frac{1}{|\mathcal{S}_{O/Y}|} \sum_{o \in \mathcal{S}_{O/Y}} h_o^L$, respectively, where $S_{O}$ is the competitive node set with the top $O$ smallest semantic distances, and $S_{O/Y}$ is the non-target node set by removing the target node set $S_{Y}$ from $S_{O}$.
We only consider the competitive node set since the average integrated semantic distance of all the nodes is significantly large and loses effectiveness in training.
With the output average semantic distances of supporting passages and non-target nodes, we formulate the hinge objective function with threshold $r$ as follows:
\begin{equation}
\setlength\abovedisplayskip{3pt}
\setlength\belowdisplayskip{3pt}
\label{gnn_object}
    \ell = \max (0, r + \bar{d}_{Y}^L - \bar{d}_{O/Y}^L).
\end{equation}
We update the GNN using the gradient descent, and training details are elaborated in Appendix \ref{appendix_gnn}.

\subsection{RGNN Enhanced Retrieval for Multi-hop Reasoning Questions}
\label{RGNN}
The information asymmetry phenomenon also frequently arises in complex multi-hop reasoning questions, as analyzed in the Introduction section.
Although recent works have incorporated multi-hop reasoning or question rewriting into retrieval \cite{yao2022react, ircot, iter_retgen}, LLMs may generate hallucinated reasons or incorrect subquestions due to the absence of prior knowledge to the question \cite{huang2023survey}.
These hallucinated reasons or incorrect subquestions fail to retrieve supporting passages when only considering the retrieval process in isolation for each step.
Additionally, they also encounter another challenge: whether to continue the next-step answering process or output the final answer.
To overcome these challenges, we first propose a \textit{self-critique} technique by prompting LLMs to determine the termination of the answering process.
To enhance retrieval of supporting passages over steps, we further utilize an RGNN to enable the integration between graphs of passages in different steps.

\textbf{Self-critique of LLMs. }
To answer multi-hop reasoning questions, \textit{self-critique} involves prompting the LLM to generate subquestions and answer them using retrieved passages in a step-by-step manner.
As depicted in Fig. \ref{fig:illustration}, at each step, after the LLM generates a subanswer to the subquestion, the question and all the generated subanswers are input into the LLM and critiqued to determine whether the generated subanswers are evident to generate the final answer to the question.
If the output is `Yes', the LLM will be prompted to generate the final answer.
Otherwise, it will be requested to generate the next-step subquestion.
Compared with \textit{SelfAsk} \cite{self_ask}, we omit the subquestions and focus on whether existing evidences are satisfactory for answering the question when determining the termination.
This \textit{self-critique} technique decomposes the question over steps and enables a more accurate answer to the question.

\textbf{Recurrent graph neural network.}
To enhance retrievals over steps, as shown in Fig. \ref{fig:illustration}, we utilize an RGNN to establish relationships between the semantic distances of subquestions.
Specifically, with the subquestion $q^t$ at step $t$, we first compute $h_i^{L,t}$ for each node $v_i \in \mathcal{V}$ using the aforementioned $L$-layer GNN.
To consider the effect of semantic distances from previous subquestions, we integrate $h_i^{L,t}$ with the previous-step integrated semantic distance to compute the integrated semantic distance $\hat{h}_i^{L,t}$ at step $t$ using the parameter $\beta$. This can be formulated as follows:
\begin{equation}
\setlength\abovedisplayskip{3pt}
\setlength\belowdisplayskip{3pt}
\label{h_hat}
    \hat{h}_i^{L,t} = \beta h_i^{L,t} + (1-\beta) \hat{h}_i^{L,t-1}, \quad (t>1),
\end{equation}
where $\hat{h}_i^{L,t-1}$ is the integrated semantic distance from the previous step $t-1$ and $\hat{h}_i^{L,1} = h_i^{L,1}$.
The integrated semantic distance $\hat{h}_i^{L,t}$ will be used to integrate with ${h}_i^{L,t+1}$ of node $v_i$ at the next step $t+1$ recurrently until the last step of the answering process $T$.
GNN facilitates the integration of semantic distances between related passages at each step, and the recurrent parameter $\beta$ enables the integration of semantic distances of different subquestions across steps, which effectively mitigates the impacts of incorrect subquestions and enhances the retrieval of supporting passages for them.

\textbf{Hinge objective for RGNN.}
We adopt a hinge objective for RGNN by considering all the $T$ answering steps for question $q$.
With the subquestion $q^t$ at step t, the primary objective of RGNN is to reduce the semantic distance of its corresponding supporting passage and make it retrieved.
Furthermore, to enhance the retrieval process for subsequent answering steps, we also reduce the semantic distances of supporting passages that should appear at subsequent steps to make them retrieved subsequently.
Specifically, at step $t$, we adopt a hinge objective that encourages the average integrated semantic distance of supporting passages that should appear at the current and subsequent steps $t+ = \{t, \cdots, T\}$ to be lower than that of the non-target node set.
We first define the average integrated semantic distances of supporting passages and non-target nodes at step $t$ as $\bar{d}_{Y^{t+}}^{L,t} = \frac{1}{|\mathcal{S}^t_{Y^{t+}}|} \sum_{j \in \mathcal{S}^t_{Y^{t+}}}\hat{h}_j^{L,t}$ and $\bar{d}_{O/Y^{t+}}^{L,t} = \frac{1}{|\mathcal{S}^t_{O/Y^{t+}}|} \sum_{o \in \mathcal{S}^t_{O/Y^{t+}}} \hat{h}_o^{L,t}$, respectively, where $\mathcal{S}_O^t$ is the competitive node set with top $O$ smallest semantic distances, and $\mathcal{S}_{O/Y^{t+}}^{t}$ is the non-target node set by removing the target node set $\mathcal{S}_{Y^{t+}}^t$ from $\mathcal{S}_O^{t}$.
The hinge objective function with threshold $r$ for the RGNN is formulated as follows:
\begin{equation}
\setlength\abovedisplayskip{3pt}
\setlength\belowdisplayskip{3pt}
    \ell = \frac{1}{T} \sum_{t \in [T]} \max (0, r + \bar{d}_{Y^{t+}}^{L,t} - \bar{d}_{O/Y^{t+}}^{L,t}).
\end{equation}
By employing this hinge objective, the RGNN takes into account the retrieval of subsequent answering steps and proactively reduces the semantic distances associated with them.
These small semantic distances are then transferred to the next subquestion and enhance the retrieval of the corresponding supporting passages.
We update the RGNN (parameters $\alpha^l$ and $\beta$) using the gradient descend, which are elaborated in Appendix \ref{appendix_rgnn}.

\section{Experiments}

In this section, we conduct experiments to demonstrate the effectiveness of \textit{GNN-Ret} and \textit{RGNN-Ret}, including \textbf{1)} superiority of \textit{GNN-Ret} and \textit{RGNN-Ret} on various QA datasets (Table \ref{main result}),
\textbf{2)} ablation studies to verify the effectiveness of components in GoPs, \textit{GNN-Ret}, and \textit{RGNN-Ret} (Table \ref{tab: ablation_component} and \ref{tab: ablation_graph}),
and \textbf{3)} superiority of \textit{GNN-Ret} on open-sourced LMs (Table \ref{tab:open_source}).
Additional analysis on hyperparameter selection, statistics of retrieval accuracy, token number for retrieval, multi-layer \textit{GNN-Ret}, extension of \textit{RGNN-Ret} to other multi-hop answering methods, and qualitative results are discussed in Appendix \ref{selection_k_o}, \ref{ret_acc}, \ref{token_number}, \ref{multi_layer_appendix}, \ref{effective_rgnn_ret}, \ref{case study}, respectively.

\subsection{Experimental Setups}

\textbf{Evaluation datasets.}
We measure all the methods on four different QA datasets, including multi-hop Wikipedia reasoning datasets: (1) MuSiQue \cite{trivedi2022musique}, (2) IIRC \cite{ferguson2020iirc}, (3) 2WikiMQA \cite{2wiki}, and a single-hop multi-choice QA dataset: (4) Quality \cite{quality}.
As evaluation metrics, we calculate the F1 score, exact match (EM) and accuracy (Acc) for multi-hop reasoning datasets.
We use the ChatGPT (\texttt{gpt-3.5-turbo-2023-06-01-preview}) to evaluate if the prediction matches with the gold answer, following \cite{iter_retgen, wang2024knowledge}.
Since Quality is not a multi-hop dataset, we only validate accuracy performance of \textit{GNN-Ret} on it.

\textbf{Baselines.}
We compare \textit{GNN-Ret} and \textit{RGNN-Ret} with the following baselines: (1) \textit{Direct} answers questions without retrieved passages.
(2)We use retrievers \textit{bm25} \cite{bm25}, \textit{DPR} \cite{karpukhin2020dense}, and \textit{SentenceBert}\footnote{\url{https://huggingface.co/sentence-transformers/multi-qa-mpnet-base-cos-v1}} (\textit{SBERT}) \cite{sbert} to retrieve passages without considering relatedness between passages for QA.
(3) \textit{SelfAsk} \cite{self_ask} prompts LLMs to generate the follow-up question and answers it with retrieved passages until generating the final answer.
(4) \textit{ITER-RETGEN} \cite{iter_retgen} iteratively answers questions with retrieved passages and uses the generated answer for the next-step retrieval until the generation of final answer.
(5) \textit{IRCoT} \cite{ircot} iteratively prompts LLMs to generate chains of thoughts with retrieved passages and retrieves with the generated reasons until reaching the maximum token number.
All the retrieved passages are then used to generate the final answer.
(6) \textit{KGP} \cite{wang2024knowledge} first searches seed passages using the question and then prompts LLMs to generate the needed evidence to retrieve the other relevant passages among neighbors of seed nodes using semantic distances.

\begin{table*}[]
\tiny
\renewcommand{\arraystretch}{0.9}
\centering
\vspace{-0.3in}
\resizebox{0.95\textwidth}{!}{
\begin{tabular}{l|cccccccccccc}
\toprule
\multirow{2}{*}{Methods} & \multicolumn{3}{c}{MuSiQue} & \multicolumn{3}{c}{IIRC} & \multicolumn{3}{c}{2WikiMQA} & \multicolumn{3}{c}{Quality}         \\ \cmidrule(lr){2-4} \cmidrule(lr){5-7} \cmidrule(lr){8-10} \cmidrule(lr){11-13}
                         & F1      & EM      & Acc     & F1     & EM     & Acc    & F1       & EM      & Acc     & F1 & EM & Acc                       \\ \midrule 
\multicolumn{13}{c}{No Retrieval}                                                                                                           \\  \midrule  
\textit{Direct}                  & 14.1    & 4.0     & 9.6     & 15.8   & 9.6    & 16.0   & 24.4     & 17.3    & 24.8    & -      & -      & 39.5 \\ \midrule 
\multicolumn{13}{c}{One-hop Retrieval}                                                                                                                 \\ \midrule 
\textit{bm25}                    & 22.2    & 10.2    & 18.8    & 27.3   & 15.4   & 31.7   & 33.8     & 23.8    & 34.6    & -      & -      & 47.0  \\
\textit{DPR}                    & 23.8    & 10.8    & 17.5    & 29.9   & 16.3   & 35.6   & 36.2     & 24.4    & 36.3    & -      & -      & 52.5  \\
\textit{SBERT}                    & 28.6    & 13.5    & 23.8    & 31.6   & 17.5   & 40.4   & 43.4     & 31.0    & 42.3    & -      & -      & 53.4  \\
\textit{GNN-Ret} (K=5)           & \textbf{31.6}    & \underline{16.5}    & \underline{27.1}    & 32.2   & 17.6   & \underline{44.0}   & \underline{47.7}     & 32.7    & 44.8    & -      & -      & \underline{55.5} \\
\textit{GNN-Ret} (K=10)          & 30.0    & 14.4    & 24.4    & \underline{35.2}   & \underline{19.6}   & \underline{44.0}   & \underline{47.7}     & \underline{32.9}    & \underline{46.3}    & -      & -      & \textbf{58.2}  \\
\textit{GNN-Ret} (w. train)   & \underline{31.1}    & \textbf{17.5}    & \textbf{27.3}    & \textbf{36.1}   & \textbf{21.9}   & \textbf{44.4}   & \textbf{48.1}     & \textbf{33.5}    & \textbf{46.9}    &    -    &     -   & -                          \\ \midrule 
\multicolumn{13}{c}{Multi-hop Retrieval}                                                                                                            \\  \midrule   \textit{SelfAsk}& 28.0    & 18.3    & 22.9    & 37.1   & 29.6   & 36.9   & 51.2     & 41.5    & 44.8    & -      & -      & -                        \\
\textit{ITER-RETGEN}           & 30.0    & 16.7    & 27.1    & 32.6   & 20.0   & 39.6   & 45.6     & 33.3    & 43.8    & -      & -      & -                       \\
\textit{IRCoT}                 & 29.9    & 12.5    & 27.5    & 32.5   & 18.8   & 41.0   & 46.0     & 30.8    & 44.6    & -      & -      & -                        \\
\textit{KGP}             & 30.0    & 15.4    & 24.0    & 33.9   & 18.5   & 42.1   & 45.5     & 31.9    & 45.2    & -      & -      & -                        \\
\textit{RGNN-Ret} & \underline{32.9}    & \underline{20.8}    & \textbf{31.3}    & \textbf{43.4}   & \textbf{28.3}   & \underline{48.1}   & \underline{59.7}     & \underline{43.3}    & \textbf{55.8}    & -      & -      & -                \\
\textit{RGNN-Ret} (w. train) & \textbf{34.8}    & \textbf{21.9}    & \textbf{31.3}    & \underline{42.8}   & \underline{27.6}   & \textbf{48.4}   & \textbf{61.4}     & \textbf{45.2}    & \underline{55.6}    & -      & -      & -                      \\ \midrule
\end{tabular}}
\vspace{-0.1in}
\caption{F1/EM/Acc for different QA methods with ChatGPT on four QA datasets. The best and second best scores are highlighted in \textbf{bold} and \underline{underline}, respectively}
\label{main result}
\vspace{-0.2in}
\end{table*}

\textbf{Implementation details.}
We use ChatGPT (\texttt{gpt-3.5-turbo-2023-06-01-preview}) as the LLM backbone for experiments and adopt a temperature of 0 to remove the effect of random sampling.
We employ SBERT as the embedding model for all our methods and baselines. 
The semantic distance is derived by $ (1- \text{cosine similarity)}$ following \cite{sarthi2023raptor}.
We set up a maximum token number of 3500 for retrieval to leave the space to instruction and demonstrations, and all the methods retrieve the semantically similar passages until reaching the maximum token number.
We adopt a one-layer GNN for our proposed methods \textit{GNN-Ret} and \textit{RGNN-Ret}.
We set up $\alpha^1=0.5$ for \textit{GNN-Ret} and $\alpha^1=0.5$ and $\beta=0.9$ for \textit{RGNN-Ret} when there is no training.
When training the RGNN, we set up $r=0.01$, $K=5$, and $O=25$ for \textit{GNN-Ret} (w. train) and $r=0.01$, $K=5$, and $O=10$ for \textit{RGNN-Ret} (w. train), respectively, which achieve the best performance in grid search experiments in Appendix \ref{selection_k_o}.
We do not train the parameters for Quality dataset without the ground-truth labels of supporting passages.
Since training the RGNN requires ground-truth subquestions, we manually generate subquestions for 5 questions sampled from the preset training data.
More implementation details are presented in Appendix \ref{appendix_details}.

\subsection{Main Results}
Table \ref{main result} summarizes the results for our proposed methods and baselines on four QA datasets using ChatGPT.
Results show that \textit{GNN-Ret} significantly outperforms baselines in terms of F1, EM, and accuracy on all these four QA datasets with a single retrieval.
For instance, compared with \textit{SBERT}, \textit{GNN-Ret} improves EM by 4.4 and accuracy by $4\%$ on the IIRC dataset. 
Additionally, \textit{GNN-Ret} maintains its superiority on the Quality dataset, which requires a comprehensive understanding of the context of an entire story.
This can be attributed to the ability of \textit{GNN-Ret} to enhance the retrieval prioritization of supporting passages through the incorporation of structural information.
Surprisingly, our proposed one-hop \textit{GNN-Ret} even slightly outperforms baselines with multi-hop retrieval processes and queries of LLMs in accuracy, highlighting significant potentials of leveraging passage relatedness to enhance the retrieval process for answering multi-hop reasoning questions.

Furthermore, results in Table \ref{main result} demonstrate that our proposed \textit{RGNN-Ret} achieves state-of-the-art performance in terms of F1, EM, and accuracy on the three multi-hop reasoning datasets. \textit{RGNN-Ret} outperforms the best baseline, \textit{KGP}, by more than $10\%$ in accuracy on 2WikiMQA. This performance improvement can be attributed to our proposed \textit{self-critique} technique and RGNN.
The \textit{self-critique} technique enables more accurate judgement on the ending of the answering process and the generation of final answers.
The incorporation of RGNN further enhances retrieval for each step of answering.

\subsection{Additional Analysis}

\textbf{Effectiveness of components in \textit{GNN-Ret}.}
To assess the individual contributions of each proposed component, we conduct ablation studies and present the results in Table \ref{tab: ablation_component}.
We explore an alternative approach by utilizing the mean semantic distance (\textit{GNN-Ret (mean)}) instead of the minimum semantic distance as the message in equation \eqref{min} for the GNN. 
This modification results in a lower accuracy compared to the use of the minimum semantic distance as the message, which suggests that employing the minimum semantic distance as the message effectively filters out interfering messages from irrelevant neighbors and preserves the most relevant message for the GNN.

\textbf{Effectiveness of components in \textit{RGNN-Ret}.}
We conduct ablation studies to evaluate the effectiveness of each component in \textit{RGNN-Ret} (i.e., recurrent, GNN, and RGNN).
Results in Table \ref{tab: ablation_component} show that both the recurrent part (parameter $\beta$) and GNN improve accuracy compared with setting with SBERT on MuSiQue and 2WikiMQA datasets.
By combining the recurrent part and GNN, our proposed method \textit{RGNN-Ret} further improves accuracy up to $31.3\%$ and $55.6\%$ on MuSiQue and 2WikiMQA datasets, respectively, which demonstrates the effectiveness of \textit{RGNN-Ret} in enhancing the retrieval by interconnecting with the GoPs from previous steps. This interconnection enhances the retrieval of supporting passages across multiple steps, leading to more accurate answers in the multi-hop reasoning process.

\begin{table}[t]
\vspace{-0.3in}
\tiny
\renewcommand{\arraystretch}{0.9}
\centering
\resizebox{0.48\textwidth}{!}{
\begin{tabular}{rcc}
\toprule
\multicolumn{1}{c}{{Methods}} & {MuSiQue}  &  2WikiMQA   \\ \hline 
{SBERT}  &  23.8 & 42.3 \\  
{GNN-Ret (mean)}  & 26.3 (+2.5) & 44.6 (+1.7) \\ 
{GNN-Ret}  &  \textbf{27.3} (+3.5) & \textbf{46.9} (+4.6) \\ \hline 
Self-critique + SBERT & 27.9 & 52.9 \\ 
Self-critique + recurrent & 28.1 (+0.2) & 54.0 (+1.1) \\ 
Self-critique + GNN &  30.8 (+2.9) & 55.2 (+2.3) \\
RGNN-Ret  &  \textbf{31.3} (+3.4)  & \textbf{55.6} (+2.7) \\ \hline 
RGNN-Ret ($Y$)  &  27.1 &  53.3\\
RGNN-Ret ($Y^t$)  & 30.2 (+3.1) & 54.6 (+1.3) \\
RGNN-Ret ($Y^{t+}$)&  \textbf{31.3} (+4.2)  & \textbf{55.6} (+2.3) \\ \toprule

\end{tabular}}
\vspace{-0.2in}
\caption{Ablation studies of components of GNN-Ret and RGNN-Ret in accuracy with ChatGPT on MuSiQue and 2WikiMQA.}
\label{tab: ablation_component}
\vspace{-0.25in}
\end{table}

We also explore different settings for the target node sets during the training of RGNN and present results in Table \ref{tab: ablation_component}. When using the entire node set of ground-truth supporting passages $\mathcal{S}_Y$ as the retrieval labels for each subquestion, denoted as \textit{RGNN-Ret} ($Y$), the accuracy decreases to $27.1\%$ and $53.3\%$ on MuSiQue and 2WikiMQA datasets, respectively. 
This is because the supporting passages for initial steps are irrelevant to the subquestions at subsequent steps. Including all of them as labels for each subquestion disrupts the training of RGNN.
Instead, we employ the corresponding label for each subquestion, denoted as \textit{RGNN-Ret} ($Y^t$). This approach effectively improves the accuracy to $30.2\%$ and $54.6\%$ on the MuSiQue and 2WikiMQA datasets, respectively. 
Futhremore, our proposed \textit{RGNN-Ret} method includes label set of supporting passages not only for the current subquestion but also for subsequent subquestions. 
This comprehensive approach further improves the accuracy to $31.3\%$ and $55.6\%$ on MuSiQue and 2WikiMQA datasets, respectively, which verifies the effectiveness of \textit{RGNN-Ret} in enhancing the retrieval process for subsequent subquestions.

\textbf{Effectiveness of GoPs.}
We explore the effectiveness of components (structural information (SI) and shared keywords (SK)) for graph construction.
Results in Table \ref{tab: ablation_graph} show that both structural information and shared keywords are able to improve accuracy for the baseline without GoPs. Notably, in the context-specific Quality story dataset, which demands a comprehensive grasp of the entire narrative for accurate retrieval, the graph that incorporates structural information markedly outperforms the setting that relies solely on shared keywords. 
While for 2WikiMQA that requires knowledge of multiple documents, the graph with shared keywords achieves a better accuracy compared with that with structural information. By combining them into graph construction, we achieve the best accuracy in these two datasets.

\begin{table}[t]
\renewcommand{\arraystretch}{0.9}
\vspace{-0.3in}
\tiny
    \centering
\resizebox{0.48\textwidth}{!}{
\begin{tabular}{rcc}
\toprule
\multicolumn{1}{c}{{Methods}} & \multicolumn{1}{c}{{2WikiMQA}}  &  {Quality}    \\   \hline
{SBERT (w.o. graph)}  &   42.3 & 53.4 \\  
{graph w. SI}  & 42.7 (+0.4) & 55.5 (+2.1) \\ 
{graph w. SK}  &  45.4 (+3.1) & 54.8 (+1.4) \\ graph w. SI and SK & \textbf{46.9} (+4.6) & \textbf{58.2} (+4.8) \\
\toprule
\end{tabular}}
\vspace{-0.2in}
\caption{Ablation studies of components on graph construction in accuracy with ChatGPT on 2WikiMQA and Quality. SI and SK represent structural information and shared keywords, respectively.}
\vspace{-0.2in}
\label{tab: ablation_graph}
\end{table}

\textbf{Performance of \textit{GNN-Ret} with open-sourced LMs.}
Our proposed \textit{GNN-Ret} utilizes the relatedness of passages to enhance retrieval with the meticulously designed GNNs, which is generalized and robust to all the LM architectures. To substantiate this, we supplement the experiments for \textit{GNN-Ret} and SBERT with two well-behaved open-sourced LMs (Qwen/Qwen2-7B-Instruct \cite{qwen2} and google/gemma-2-9b-it \cite{gemma_2024}) from Open LLM Leaderboard \cite{open-llm-leaderboard-v2}. Results in Table \ref{tab:open_source} show that \textit{GNN-Ret} consistently outperforms SBERT across varying open-sourced LMs, demonstrating the generalization of \textit{GNN-Ret} in enhancing retrieval coverage with varying LMs.

\begin{table}[h]
\vspace{-0.1in}
\renewcommand{\arraystretch}{0.8}
    \centering
    \resizebox{0.48\textwidth}{!}{
    \begin{tabular}{ccccccc}
    \toprule
   & \multicolumn{3}{c}{MuSiQue} &\multicolumn{3}{c}{2WikiMQA} \\ \midrule 
      Qwen2-7B-Instruct & F1 & EM & Acc & F1 & EM   & Acc \\ \midrule
      SBERT & 27.8 & 16.9 & 23.3 & 43.3 & 34.0 & 40.8 \\
      GNN-Ret & \textbf{32.3} & \textbf{20.8} & \textbf{26.7} & \textbf{50.5} & \textbf{37.3} & \textbf{45.8} \\ \midrule
      gemma-2-9b-it &  F1 & EM & Acc & F1 & EM   & Acc \\ \midrule
      SBERT & 32.8 & 18.9 & 26.9 & 46.4 & 28.8 & 43.8 \\
      GNN-Ret & \textbf{34.6} & \textbf{20.8} & \textbf{29.0} & \textbf{51.6} & \textbf{30.8} & \textbf{45.8} \\
      \toprule
    \end{tabular}}
    \vspace{-0.15in}
    \caption{F1/EM/Acc of SBERT and GNN-Ret with open-sourced LMs on MuSiQue and 2WikiMQA.}
    \vspace{-0.2in}
    \label{tab:open_source}
\end{table}

\section{Conclusion}

In this paper, we proposed \textit{GNN-Ret}, an effective method to enhance retrieval for QA of LLMs by exploiting the inherent relatedness between passages on a graph of passages.
By extending \textit{GNN-Ret} to multi-hop reasoning questions, we proposed \textit{RGNN-Ret}, which enhances the retrieval for subquestions through the interconnection between graphs of passages across steps.
The experiments clearly demonstrated the superiority of both \textit{GNN-Ret} and \textit{RGNN-Ret} over baselines, highlighting the effectiveness of leveraging relatedness between passages to enhance retrieval.
From these advantages, we believe that the incorporation of graph-based representations and the exploitation of passage relatedness can open up new avenues of research in the field of LLMs in understanding structural documents and answering complex questions.

\section{{Limitations}}
\label{limitations}

\textbf{Costs of graph construction.}
While our proposed \textit{GNN-Ret} demonstrates impressive accuracy in QA tasks with a single query of LLMs, it still relies on multiple LLM queries to extract keywords from passages and construct the graph of passages. 
Fortunately, there are alternative fine-tuned language models specific for keyword extraction\footnote{\url{https://huggingface.co/ml6team/keyphrase-extraction-distilbert-inspec}}\footnote{\url{https://github.com/MaartenGr/KeyBERT}}.
These models are more compact and significantly accelerate the graph construction process.
Moreover, the flexibility of the graph of passages allows for dynamic modifications, such as adding new passage nodes or removing outdated ones. 
By maintaining a domain-specific graph of passages, we can effectively address various questions for this domain without the need for reconstructing the graph.

\textbf{Costs of message passing.}
We first analyze the statistics of the graph of passages in experiments and detail results in Table \ref{statistics_graph}. 
In larger datasets, each node typically exhibits a higher number of neighbors, attributed to an increased likelihood of keyword sharing among more nodes. We calculate the density of the graph of passages, denoted as $D=\frac{2|\mathcal{E}|}{|\mathcal{V}|(|\mathcal{V}|-1)}$, for each dataset in our experiments, based on the established definition \cite{density_graph}, where $\mathcal{E}$ and $\mathcal{V}$ represent the edges and vertices of the graph, respectively. The densities of all graphs are significantly less than 1, underscoring the substantial sparsity observed in these graphs across our experimental datasets.

We adopt relevant nodes sampling (Section \ref{GNN}) to allow the relevant node set with only top $K$ smallest semantic distances to pass the information to their neighbors during message passing. Assuming an average number of neighbors $\mathcal{V}_K$ among these $K$ relevant nodes, the time complexity for message passing can be derived as $O(K|\mathcal{V}_K|)$, considering a single-layer GNN. 
With $K$ set to 5 or 10 in our experiments, and given the small empirical average number of edges per node $\frac{|\mathcal{E}|}{|\mathcal{V}|}$ reported in Table \ref{statistics_graph}, the time complexity of message passing is neglectable compared with LLM queries.

\begin{table}[]
\centering
\resizebox{0.48\textwidth}{!}{
\begin{tabular}{cccc}
\toprule
Datasets & Number of nodes ($|\mathcal{V}|$) & Average number of edges ($\frac{|\mathcal{E}|}{|\mathcal{V}|}$) & Density \\ \midrule 
2WikiMQA & 9815            & 91.38                   & 0.018   \\
IIRC     & 21866           & 259.22                  & 0.024   \\
MuSiQue  & 20071           & 335.77                  & 0.033   \\
Quality  & 1104            & 26.00                   & 0.047   \\
\toprule
\end{tabular}}
\caption{Statistics of graph of passages in experiments.}
\label{statistics_graph}
\end{table}

\textbf{Undirected Graph.}
In this study, our focus lies solely on improving the retrieval process through an undirected graph of passages, with no additional information associated with the edges. While we employ a sampling technique to select relevant nodes and utilize the minimum semantic distance as the message in the GNN, we acknowledge that there is still a possibility of irrelevant passages receiving messages, potentially impacting the retrieval performance. Consequently, the task of selecting an appropriate path on the graph of passages that aligns with the given question remains an area that is yet to be thoroughly explored. 
In addition, we only consider the structural-related passages that are in the same section.
However, some questions may require more complex structural information.
For example, when the question is about comparison of two entities in different sections, more complex structural information is needed to connect these corresponding passages to enhance retrieval.
Therefore, a more delicate graph of passages for more complex tasks warrant further investigation in future studies.

\bibliography{custom}

\appendix

\section{Training Details}
\label{train_details}

In this section, we supplement the training details of GNN and RGNN as below.

\subsection{Training of GNN}
\label{appendix_gnn}

\textbf{Hinge objective for GNN.}
Recall that the objective of GNN is to reduce the semantic distances of supporting passages more than those of non-target nodes.
With the ground-truth set of supporting passages $\mathcal{S}_Y$, we first define the average integrated semantic distances of supporting passages $\bar{d}_{Y}^L$ and non-target nodes $\bar{d}_{O/Y}^L$ as below:
\begin{align}
    \bar{d}_{Y}^L = \frac{1}{|\mathcal{S}_{Y}|} \sum_{j \in \mathcal{S}_Y} h_j^L, \\
    \bar{d}_{O/Y}^L = \frac{1}{|\mathcal{S}_{O/Y}|} \sum_{o \in \mathcal{S}_{O/Y}} h_o^L,
\end{align}
where $S_{O}$ is the competitive node set with the top $O$ smallest semantic distances and $S_{O/Y}$ is the non-target node set by removing the target node set $S_{Y}$ from $S_{O}$, respectively.
We only consider the competitive node set since the average integrated semantic distance of all the nodes is significantly large and loses effectiveness in training.
With the output average semantic distances of supporting passages and non-target nodes, we formulate the hinge objective function with threshold $r$ as follows:
\begin{align}
    \ell = \max (0, r + \bar{d}_{Y}^L - \bar{d}_{O/Y}^L).
\end{align}

We update the parameters of GNN using gradient descent, and the backward process is presented as below.
With only partial nodes accepting messages from neighbors, we first define the received node set at each layer $l$ as follows: 
\begin{align}
    \mathcal{S}_R^l = \left\{
    \begin{aligned}
        \mathcal{S}_O \cap \mathcal{S}_K^L, \quad l=L \\
        \mathcal{S}_K^l, \quad \text{else}
    \end{aligned}
    \right. .
\end{align}
For each node $i \in \mathcal{S}_R^L$ at the last layer $L$, the gradient of $\ell$ w.r.t. $h_i^L$ is:
\begin{align}
    \frac{\partial  \ell}{\partial h^L_i} = \left\{
    \begin{aligned}
        1.0, \quad i \in \mathcal{S}_Y \\
        -1.0 , \quad \text{else}
    \end{aligned}
    \right. .
\end{align}
For the layer $l<L$, the gradient of $l$ w.r.t. $\alpha^l$ is presented as follows:
\begin{align}
    \frac{\partial  \ell}{\partial  \alpha^l} = \sum_{j \in \mathcal{S}_R^{l}} \frac{\partial  \ell}{\partial h^l_i} (h_i^{l-1} - m_i^l).
\label{gradient_alpha}
\end{align}
Assuming that node $i \in \mathcal{T}_k^{l-1}$ at layer $l-1$ transfers the message to the node set $\mathcal{S}_{K,i}^l$ at layer $l$ (one node can transfers messages to multiple neighbors), we update the gradient of $\alpha^l$ w.r.t. $h^{l-1}_i$ at the previous layer $l-1$ as follows:
\begin{align}
    \frac{\partial  \ell}{\partial h^{l-1}_i} = \sum_{j \in \mathcal{S}_{K,i }^l} - \alpha^{l} \frac{\partial  \ell}{\partial  h_j^l} + \mathbb{1} (i \in \mathcal{S}_{K,i}^l) \alpha^l \frac{\partial  \ell}{\partial  h_j^l} .
\label{gradient_h}
\end{align}
By propagating the gradients backward to the first layer, we can obtain the gradients for each layer of $\alpha$.

\subsection{Training of RGNN}
\label{appendix_rgnn}

\textbf{Hinge objective for RGNN.}
For each subquestion $q^t$ at step $t$, we input its semantic distances into GNN and compute hidden state $h_j^{l,t}$ and message $m_j^{l,t}$ of the $j-$th node for $j \in \{1,\cdots,|\mathcal{V}|\}$ at layer $l$.
Recall that we adopt a hinge objective that encourages the average integrated semantic distance of supporting passages that should appear at the current and subsequent steps $t+ = \{t, \cdots, T\}$ to be lower than that of the non-target node set for subquestion $q^t$ at step $t$.
We first define the average semantic distances of supporting passages ${\bar{d}_{Y^{t+}}^{L,t}}$ and non-target nodes $\bar{d}_{O/Y^{t+}}^{L,t}$ at step $t$ as below:
\begin{align}
    \bar{d}_{Y^{t+}}^{L,t} = \frac{1}{|\mathcal{S}^t_{Y^{t+}}|} \sum_{j \in \mathcal{S}^t_{Y^{t+}}} \hat{h}_j^{L,t}, \\
    \bar{d}_{O/Y^{t+}}^{L,t} = \frac{1}{|\mathcal{S}^t_{O/Y^{t+}}|} \sum_{o \in \mathcal{S}^t_{O/Y^{t+}}} \hat{h}_o^{L,t},
\end{align}
where $\mathcal{S}_O^t$ is the competitive node set with top $O$ smallest semantic distances at step $t$ and  $\mathcal{S}_{O/Y^{t+}}^{t}$ is the non-target node set by removing the target node set $\mathcal{S}_{Y^{t+}}^t$ from $\mathcal{S}_O^{t}$, respectively.
We formulate the hinge objective function with threshold $r$ for the RGNN as follows:
\begin{align}
    \ell &= \frac{1}{T} \sum_{t \in [T]} \ell^t \nonumber \\
    &= \frac{1}{T} \sum_{t \in [T]} \max (0, r + \bar{d}_{Y^{t+}}^{L,t} - \bar{d}_{O/Y^{t+}}^{L,t}).
\end{align}
We update the parameters of RGNN using gradient descent, and the backward process is presented as below. With only partial nodes accepting messages from neighbors, we first define the recieved node set $\mathcal{S}_R^{l,t}$ at layer $l$ at step $t$ as follows:
\begin{align}
    \mathcal{S}_R^{l,t} = \left\{
    \begin{aligned}
        \mathcal{S}_O^t \cap \mathcal{S}_K^{L,t}, \quad l=L \\
        \mathcal{S}_K^{l,t}, \quad \text{else}
    \end{aligned}
    \right. .
\end{align}
For each node $i \in \mathcal{S}_R^{L,t}$ at the last layer $L$, the gradient of $\ell^t$ w.r.t. $\hat{h}_i^{L,t}$ is:
\begin{align}
    \frac{\partial  \ell^t}{\partial \hat{h}^{L,t}_i} = \left\{
    \begin{aligned}
        1.0, \quad i \in \mathcal{S}_{Y^t}^t \\
        -1.0 , \quad \text{else}
    \end{aligned}
    \right. .
\end{align} 
Therefore, the gradient of $\ell$ w.r.t. $\hat{h}_i^{L,t}$ is:
\begin{align}
    \frac{\partial  \ell}{\partial \hat{h}^{L,t}_i} = \sum_{\tau=t}^T \frac{\partial  \ell^\tau}{\partial \hat{h}^{L,t}_i} = \frac{\partial  \ell^t}{\partial \hat{h}^{L,t}_i} + \frac{\partial  \ell}{\partial \hat{h}^{L,t+1}_i} (1-\beta),
\end{align}
where 
\begin{align}
    \frac{\partial  \ell}{\partial \hat{h}^{L,T}_i} =  \frac{\partial  \ell^{T}}{\partial \hat{h}^{L,T}_i} .
\end{align}

Each node $j \in \mathcal{V}$ at step $t$ transfers the message to the same node at the next step through $\beta$. Therefore, the gradient of loss $\ell$ with respect to $\beta$ is computed by:
\begin{align}
    \frac{\partial \ell}{\partial \beta} = \frac{1}{T} \sum_{t=1}^T \frac{1}{|\mathcal{V}|} \sum_{j \in \mathcal{V}} \frac{\partial \ell}{\partial \hat{h}^{L,t}_j} (h_j^{L,t}-\hat{h}_j^{L,t-1}).
\end{align}
Recall that $\hat{h}^{L,t}_i = \beta {h}^{L,t}_i + (1-\beta){h}^{L,t-1}_i $ for each node $i \in \mathcal{V}$.
The gradient of loss $\ell$ with respect to ${h}^{L,t}_i$ is computed by:
\begin{align}
    \frac{\partial  \ell}{\partial {h}^{L,t}_i} = \beta \frac{\partial  \ell}{\partial \hat{h}^{L,t}_i}.
\end{align}
With $\frac{\partial  \ell}{\partial {h}^{L,t}_i}$, we propagate the gradient backward to the GNN and update the parameter $\alpha^l$ layer by layer using \eqref{gradient_alpha} and \eqref{gradient_h}.

\section{Additional Experimental Details}
\label{appendix_details}

\textbf{Datasets processing.}
We measure all the methods on four different QA datasets, including multi-hop Wikipedia reasoning datasets: (1) MuSiQue \cite{trivedi2022musique}, (2) IIRC \cite{ferguson2020iirc}, (3) 2WikiMQA \cite{2wiki}, and a single-hop multi-choice QA dataset: (4) Quality \cite{quality}.
For each multi-hop dataset, we randomly sample 500 questions from the development set, which provides the ground-truth supporting passages.
We use 20 of them to train the GNN and select the rest as test data.
We use all the Wikipedia documents of these 500 questions as the retrieval corpus, which is significantly larger than that provided by datasets\footnote{Prior works often use 10–20 paragraphs or documents as the retrieval corpus for each question \cite{iter_retgen, wang2024knowledge}.}.
For Quality, we randomly sample 30 articles with overall corresponding 561 questions and use these articles as the retrieval corpus.
For each document or article, we split it into multiple passages with a maximum token length of 200 \cite{sarthi2023raptor}.

\textbf{Graph of passages construction.}
We constructe a large graph of passages to serve as the retrieval corpus for each evaluation dataset.
For the Quality dataset, we randomly sample 30 articles with a diverse range of topics as the retrieval corpus. 
For multi-hop reasoning datasets, we first collect the Wikipedia documents using the provided titles required for answering these 500 questions.
For the multi-hop reasoning datasets, we collect Wikipedia documents based on the provided titles required to answer the 500 questions. 
We chunk the documents into smaller passages and record their sequential order. 
Passages that are physically adjacent to each other are regarded as structure-related and connected together in the graph.
Besides, to find out the keyword-related passages, we also extract the keywords for each passage and then connect those that share the same keywords.
Specifically, we prompt ChatGPT (\texttt{gpt-3.5-turbo-2023-06-01-preview}) to extract the Wikipedia keywords from the passages and generate their corresponding links.
The links are used to ensure the consistency of the Wikipedia document that use different keywords in passages.
The prompt for keyword extraction is given as follows:

\begin{minipage}{0.9\columnwidth}
    \centering
    \begin{tcolorbox}[title= {Prompt for extracting keywords }]
        \small
        Instruction: Extract the entities exist in this text and then provide the wikipedia links for the entities exist in this text. Output the entities and their wikipedia links in the list format, e.g., [['entity1', 'link1'], ['entity2', 'link2']].
        \\
        
         \texttt{<Passage>}
    \end{tcolorbox}
    \vspace{1mm}
\end{minipage}

The passages share the same links of keywords are considered as keyword-related and connected together in the graph.

\begin{figure*}[h]
\vspace{-0.2in}
\begin{center}
    {
    \subfigure[GNN-Ret]
        {
        \includegraphics[width=0.48\textwidth]{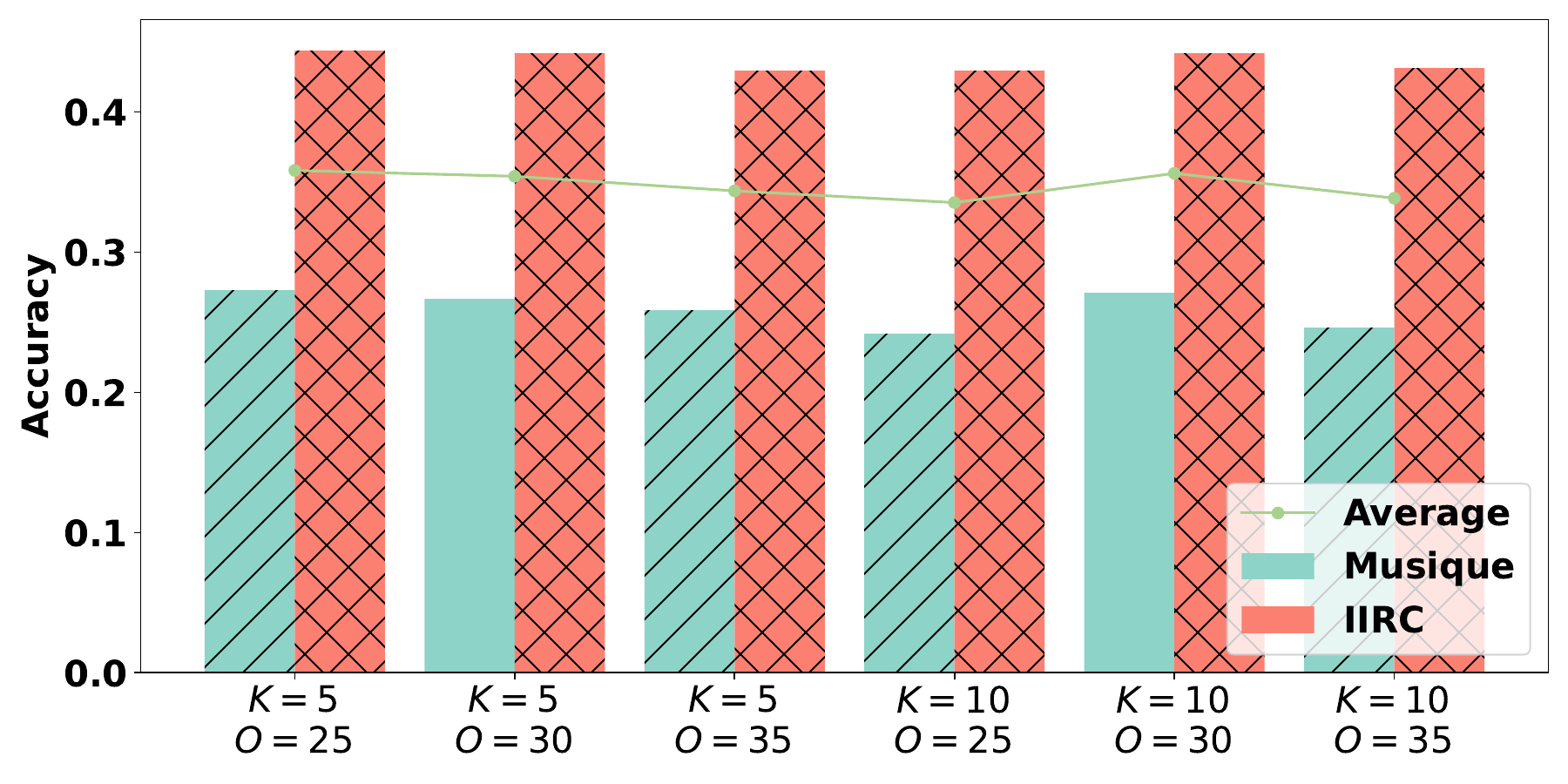}
        \label{fig:fig3a}
        }
    \subfigure[RGNN-Ret]
        {
        \includegraphics[width=0.48\textwidth]{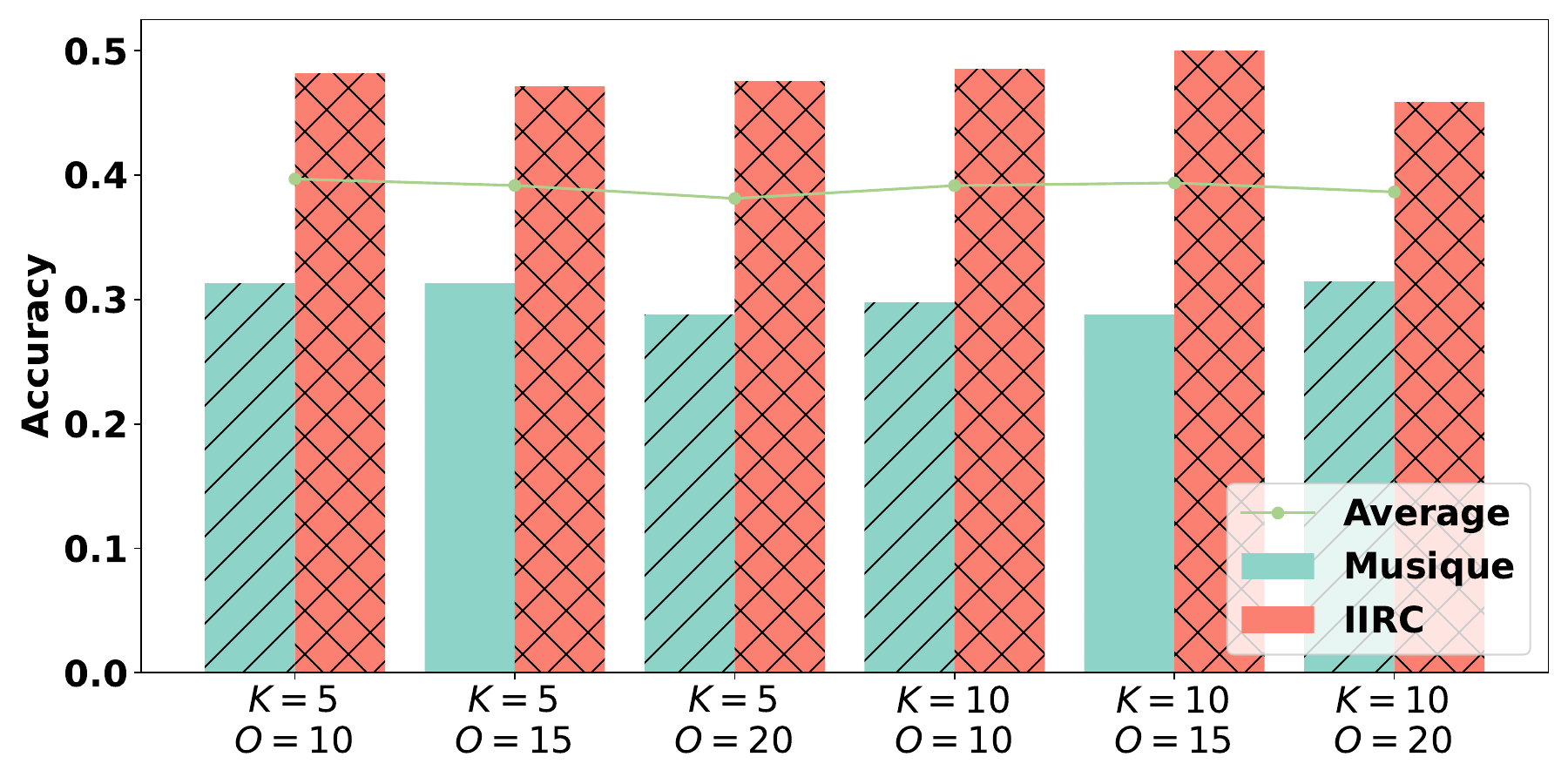}
        \label{fig:fig3b}
        } 
    }
\end{center}
\vspace{-0.2in}
\caption{Accuracy of \textit{GNN-Ret} and \textit{RGNN-Ret} with various $K$ and $O$ in Musique and IIRC datasets. The average accuracy of two datasets are displayed in {\color{green} green} points.}
\vspace{-0.2in}
\label{fig:ablation}
\end{figure*}

\textbf{Implemental details of training.}
We initialize $\alpha=0.1$ for GNN-Ret and $\alpha^1=1.0$ and $\beta=1.0$ for RGNN-Ret, respectively. We then use gradient descent to update the GNN and RGNN.
For each iteration, we compute the average gradient of all the training samples and update the parameters with a learning rate of 1.0.
The training will stop when the absolute value of gradient is smaller than 0.001 or the loss keeps increasing for five consecutive iterations.
As there is often the whole label set of supporting passages for the question but not the individual index for each subquestion, we select the one with the lowest semantic distance as the label for each subquestion and then remove this index label from the label set for subsequent subquestions.
Since training the RGNN requires ground-truth subquestions, we manually generate subquestions for 5 questions sampled from the preset training data.
The training samples of MuSiQue, IIRC, and 2WikiMQA datasets for the RGNN are displayed in Appendix \ref{train data}.
We also display the instructions and prompt templates of all the methods in Appendix \ref{prompt}.

\section{Selection of $K$ and $O$.}
\label{selection_k_o}

In order to effectively train the GNN and RGNN models, we conduct grid search for hyperparameters $K$ and $O$, which determine the number of relevant nodes to be sampled and the size of the competitive node set, respectively. 
To assess the impact of different values of $K$ and $O$, we evaluate the accuracy of \textit{GNN-Ret} and \textit{RGNN-Ret} on the MuSiQue and IIRC datasets with varying $K$ and $O$. The average accuracy across these two datasets is represented by the {\color{green}green} points in Fig. \ref{fig:ablation}. Results demonstrate that both \textit{GNN-Ret} and \textit{RGNN-Ret} consistently achieve stable accuracy performance across different settings of $K$ and $O$.
Upon analyzing the results, we observe that \textit{GNN-Ret} achieves the highest accuracy when $K=5$ and $O=25$, while \textit{RGNN-Ret} performs best with $K=5$ and $O=10$. 
\textit{GNN-Ret} adopts a larger value of $O$ since it considers the whole label set during training, while \textit{RGNN-Ret} only considers the labels for the current step and subsequent steps.
Therefore, we select these settings of $K$ and $O$ to train the GNN and RGNN models for experiments reported in Table \ref{main result}.

\begin{figure*}[t]
\begin{center}
    {
    \subfigure[MuSiQue]
        {
        \includegraphics[width=0.48\textwidth]{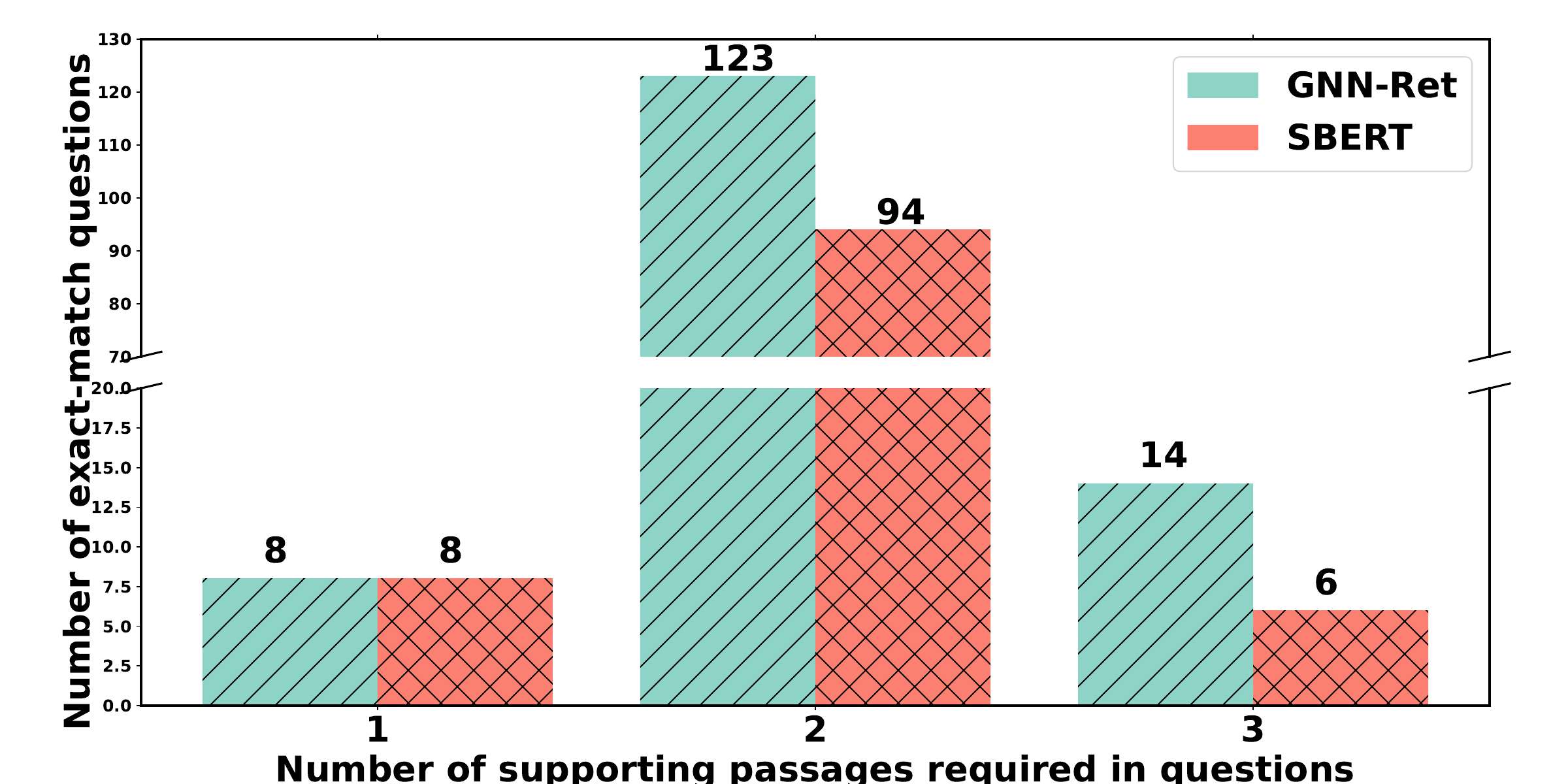}
        \label{fig:stat_iirc}
        }
    \subfigure[2WikiMQA]
        {
        \includegraphics[width=0.48\textwidth]{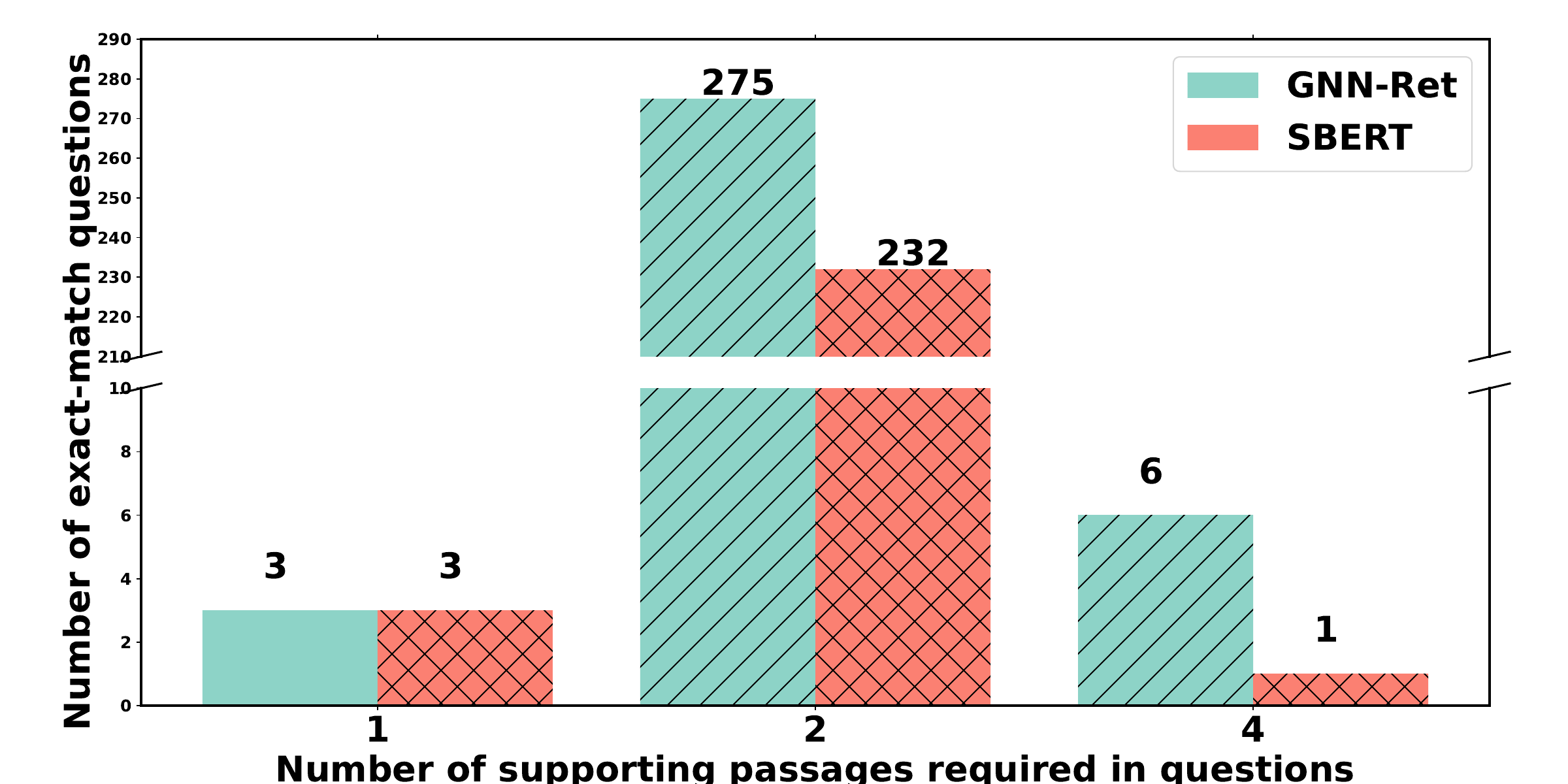}
        \label{fig:stat_2wiki}
        } 
    }
\end{center}
\vspace{-0.2in}
\caption{Exact-match number of test samples with varying numbers of supporting passages required for QA on MuSiQue and 2WikiMQA datasets.}
\vspace{-0.2in}
\label{fig:stat}
\end{figure*}

\section{Statistics of retrieval accuracy.}
\label{ret_acc}
We explore the EM between the retrieved passages and the ground-truth passages and collect the exact-match number of test samples with varying numbers of supporting passages required in questions on MuSiQue and 2WikiMQA datasets.
We do not evaluate the statistics of retrieval accuracy for \textit{RGNN-Ret} without the specific ground-truth supporting passages for each subquestion.
Results in Fig. \ref{fig:stat} show that our proposed \textit{GNN-Ret} achieves higher retrieval accuracy compared with \textit{SBERT} on MuSiQue and 2WikiMQA datasets.
For instance, \textit{GNN-Ret} outperforms \textit{SBERT} by 29 and 43 exact-match test samples for the questions that require 2 supporting passages.
This demonstrates that the GNN indeed improves retrieval coverage of supporting passages and the retrieval prioritization of supporting passages for subsequent reasoning.

\begin{table}[h]
    \centering
    \resizebox{0.48\textwidth}{!}{
    \begin{tabular}{cccccc}
    \toprule
     Token number for retrieval    &  1k & 3k & 5k & 10k& 20k \\ \midrule 
     SBERT & 18.3 & 23.8 & 22.5 & 22.5 & 27.9 \\
      GNN-Ret    &  \textbf{20.4} & \textbf{25.8} & \textbf{25.8} & \textbf{26.7} & \textbf{31.9} \\
      \toprule
    \end{tabular}}
    \caption{Accuracy of GNN-Ret and SBERT with varying token numbers for retrieval using Qwen/Qwen2-7B-Instruct on MuSiQue dataset.}
    \label{tab:token_ret}
\end{table}

\section{Ablation Studies of Token Number for Retrieval}
\label{token_number}

We conduct experiments with varying numbers of tokens for retrieval on \textit{SBERT} and \textit{GNN-Ret} with the long-context model Qwen/Qwen2-7B-Instruct \cite{qwen2}, which can handle more than 100k tokens. Results in Table \ref{tab:token_ret} show that our proposed \textit{GNN-Ret} consistently outperforms \textit{SBERT} with varying numbers of tokens for retrieval, demonstrating the robustness of our method for long contexts.

\begin{table*}[]
\centering
\resizebox{0.98\textwidth}{!}{
\begin{tabular}{ccccccccc}
\toprule
\multirow{2}{*}{GNN-Ret} & \multirow{2}{*}{Learned values of $\alpha$} & \multicolumn{3}{c}{MuSiQue} & \multirow{2}{*}{Learned values of $\alpha$} & \multicolumn{3}{c}{2WikiMQA} \\ \cmidrule(lr){3-5} \cmidrule(lr){7-9}
                         &                                             & F1      & EM      & Acc     &                                             & F1       & EM      & Acc     \\ \midrule \midrule
1 layer                  & $\alpha=[0.326]$                            & 31.1    & 17.5    & 27.3    & $\alpha=[0.277]$                            & 48.1     & 33.5    & 46.9    \\
2 layers                 & $\alpha=[0.564, 0.821]$                     & 31.5    & 16.3    & 27.3    & $\alpha=[0.582, 0.578]$                     & 47.0     & 33.3    & 46.5    \\
3 layers                 & $\alpha=[0.506, 0.511, 0.944]$              & 30.8    & 16.3    & 26.7    & $\alpha=[0.504, 0.597, 0.802]$              & 48.6     & 34.4    & 47.5   \\ \toprule
\end{tabular}}
\caption{F1, EM, and Accuracy of GNN-Ret with varying layers on MuSiQue and 2WikiMQA datasets.}
\label{multi_layers}
\end{table*}

\section{Extending GNN-Ret to Multiple Layers}
\label{multi_layer_appendix}

Our proposed GNNs are designed to be extensible across multiple layers. To explore this capability, we conducted additional experiments by increasing the number of GNN layers. The results, along with the final learned values of $\alpha$, are presented in Table \ref{multi_layers}. The results indicate that \textit{GNN-Ret} maintains consistent performance across different layer configurations. This observation could be attributed to the fact that many questions in datasets require only two-hop supporting passages, suggesting that a single-layer GNN is adequately equipped to address these queries.

\begin{table}[h]
\centering
\vspace{-0.1in}
\resizebox{0.48\textwidth}{!}{
\begin{tabular}{rcc}
\toprule
\multicolumn{1}{c}{\textbf{Method}} & \multicolumn{1}{c}{\textbf{MuSiQue}}  &  \textbf{2WikiMQA}    \\ \midrule  \midrule
IRCoT  &   27.5 & 44.6 \\  
IRCoT + RGNN-Ret  & 30.4 (+2.9) & 47.5 (+2.9) \\ 
ITER-RETGEN  &  27.1 & 43.8 \\ 
ITER-RETGEN + RGNN-Ret & 29.8 (+2.7) & 47.5 (+3.7) \\
\toprule
\end{tabular}}
\caption{Accuracy of IRCoT (w. RGNN-Ret) and ITER-RETGEN (w. RGNN-Ret) on MuSiQue and 2WikiMQA datasets.}
\label{tab: rgnn_ret_ircot}
\end{table}

\section{Effectiveness of RGNN-Ret for other multi-hop answering methods}
\label{effective_rgnn_ret}
To further demonstrate the effectiveness of \textit{RGNN-Ret}, we supplement experiments by employing it with other multi-hop baselines \textit{IRCoT} and \textit{ITER-RETGEN}. These methods utilize the generated reasons for retrieval, followed by the generation of the next-step reason or the final answer. Results in Table \ref{tab: rgnn_ret_ircot} show that \textit{RGNN-Ret} consistently and significantly improves the accuracy performance for other multi-hop baselines, which demonstrates the effectiveness and adaptability of \textit{RGNN-Ret} in enhancing the retrieval performance for multi-hop answering processes.

\section{Qualitative Results}
\label{case study}

We analyze the qualitative results in experiments and demonstrate the effectiveness of \textit{GNN-Ret} and \textit{RGNN-Ret} in improving accuracy for QA in this section.

\textbf{\textit{GNN-Ret} improves the retrieval coverage of supporting passages.}
We compare the retrieval process between \textit{SBERT} and \textit{GNN-Ret} and display the results on Table \ref{tab:case_gnn}.
\textit{GNN-Ret} can retrieve all of the supporting passages while \textit{SBERT} fails in both of these two cases.
For the first question, \textit{SBERT} can retrieve the knowledge that `\textit{the director of file Hotel By The Hour is Rolf Olsen}.'.
Since it considers passages in isolation during retrieval, the supporting passage for the inquiry information (`\textit{which country is Rolf Olsen from}') cannot be retrieved with a poor semantic distance.
In contrast, \textit{GNN-Ret} is able to retrieve both of the supporting passages about `the director of file Hotel By The Hour is Rolf Olsen' and `Rolf Olsen was an Austrian actor'.
Consequently, it can output the correct final answer.
This is attributed to the fact that \textit{GNN-Ret} takes relatedness between passages into account and thus allows the supporting passages for inquiry information to accept the effect of semantic distances from those for background information, thereby improving the retrieval coverage of supporting passages.

\begingroup
\setlength{\tabcolsep}{3pt} 
\renewcommand{\arraystretch}{1} 
\begin{table*}[t!]
    \small
    \centering
    \adjustbox{max width=0.99\textwidth}{
    \begin{tabular}{p{\textwidth}}
        \toprule
        \textbf{Question:}  Which country the director of film Hotel By The Hour is from? \\
        \textbf{Ground-truth answer:} Austria \\ \midrule
        \multicolumn{1}{c}{\textbf{\textit{SBERT}}} \\ \midrule
        \textbf{Retrieved passages:} \\
        (1) Hotel by the Hour (German title:) is a 1970 West German crime film directed by {\color{violet}Rolf Olsen} and starring Curd Jürgens, Andrea Rau and Corny Collins. \\
        (2) ... \\
        \textbf{Final answer:}
        Rolf Olsen is from {\color{red}Germany}. \\ \midrule
        \multicolumn{1}{c}{\textbf{\textit{GNN-Ret}}} \\ \midrule
        \textbf{Retrieved passages:} \\
        (1) Hotel by the Hour (German title:) is a 1970 West German crime film directed by {\color{violet}Rolf Olsen} and starring Curd Jürgens, Andrea Rau and Corny Collins. \\
        (2) {\color{violet}Rolf Olsen} (26 December 1919 – 3 April 1998) was an Austrian actor. \\
        \textbf{Final answer:} Rolf Olsen is from {\color{green}Austria}. \\
        \midrule \\
        \midrule
        \textbf{Question:} What is the date of death of the performer of song Goodbye Pork Pie Hat?\\
        \textbf{Ground-truth answer:} January 5, 1979\\ \midrule
        \multicolumn{1}{c}{\textbf{\textit{SBERT}}} \\
        \midrule
        \textbf{Retrieved passages:} \\
        (1) `Goodbye Pork Pie Hat' is a jazz standard composed by {\color{violet}Charles Mingus}, originally recorded by his sextet in 1959 as listed below, and released on his album. \\
        (2) ... \\
        \textbf{Final Answer:} {\color{red}February 17, 2017} \\ \midrule
        \multicolumn{1}{c}{\textbf{\textit{GNN-Ret}}} \\ \midrule
        \textbf{Retrieved passages:} \\
        (1)`Goodbye Pork Pie Hat' is a jazz standard composed by {\color{violet}Charles Mingus}, originally recorded by his sextet in 1959 as listed below, and released on his album. \\
        (2) {\color{violet}Charles Mingus Jr.} (April 22, 1922 – January 5, 1979) was an American jazz double bassist, pianist, composer and bandleader. \\
        \textbf{Final answer:} {\color{green}January 5, 1979} \\
        \bottomrule
    \end{tabular}
    }
    \caption{
    Two examples demonstrating effectiveness of \textit{GNN-Ret}.
    We highlight the {\color{violet}shared keywords}, {\color{green}correct answers}, and {\color{red}incorrect answers} in violet, green, and red.
    }
    \label{tab:case_gnn}
\end{table*}
\endgroup

\begingroup
\setlength{\tabcolsep}{3pt} 
\renewcommand{\arraystretch}{1} 
\begin{table*}[t!]
    \small
    \centering
    \adjustbox{max width=0.99\textwidth}{
    \begin{tabular}{p{\textwidth}}
        \toprule
        \textbf{Question:} Which award the performer of song One More Time (Joe Jackson Song) earned?\\
        \textbf{Ground-truth answer:} Grammy  \\ \midrule
        \multicolumn{1}{c}{\textbf{\textit{SelfAsk}}} \\ \midrule
        \textbf{Generation:} \\
        Here is the question:\\
Question: Which award the performer of song One More Time (Joe Jackson Song) earned?\\
Are follow up questions needed here: Yes.\\
Follow up: Who is the performer of the song One More Time (Joe Jackson Song)?\\
The performer of the song ""One More Time"" (Joe Jackson Song) is Joe Jackson.\\
So the final answer to the Question is: {\color{red}Joe Jackson}
 \\ \midrule
        \multicolumn{1}{c}{\textbf{\textit{RGNN-Ret}}} \\ \midrule
        \textbf{Generation:} \\
       Question: Which award the performer of song One More Time (Joe Jackson Song) earned? \\
Next step: What is the name of the performer of the song ""One More Time (Joe Jackson Song)""?\\
Intermediate Answer: The performer of the song ""One More Time"" is Joe Jackson.\\
Next step: What award did Joe Jackson earn for the song ""One More Time""?\\
Intermediate Answer: Joe Jackson earned a {\color{green} Grammy} for Best Pop Instrumental Album in 2001 for his album Symphony No. 1.\\
Final answer: {\color{green} Grammy}
 \\
        \midrule \\
        \midrule
        \textbf{Question:}  What other notable work did the creator of Shrek make?\\
        \textbf{Ground-truth answer:} Doctor De Soto\\ \midrule
        \multicolumn{1}{c}{\textbf{\textit{SelfAsk}}} \\
        \midrule
        \textbf{Generation:} \\
        Question: What other notable work did the creator of Shrek make?\\
Are follow up questions needed here: Yes.\\
Follow up: Who is the creator of Shrek?\\
The creator of Shrek is William Steig.\\
Are follow up questions needed here: Yes.\\
Follow up: What other notable work did William Steig make?\\
William Steig also wrote {\color{red}"Sylvester and the Magic Pebble."}\\
So the final answer is: {\color{red}Sylvester and the Magic Pebble}
 \\
         \midrule
        \multicolumn{1}{c}{\textbf{\textit{RGNN-Ret}}} \\ \midrule
        \textbf{Generation:} \\
        Question: What other notable work did the creator of Shrek make?\\
Next step: Who is the creator of Shrek?\\
Intermediate Answer: The creator of Shrek is American author William Steig.\\
Next step: What other notable works did William Steig create besides Shrek?\\
Intermediate Answer: William Steig, the creator of Shrek, also wrote and illustrated the children's picture book {\color{green}"Doctor De Soto"} which was first published in 1982...\\
Final answer: {\color{green}Doctor De Soto}
\\
        \bottomrule
    \end{tabular}
    }
    \caption{
    Two generation examples of comparison between \textit{RGNN-Ret} and \textit{SelfAsk}.
    We highlight the {\color{green}correct answers} and {\color{red}incorrect answers} in green and red. We omit the retrieved passages in the table.
    }
    \label{tab:case_rgnn}
\end{table*}
\endgroup

\textbf{\textit{RGNN-Ret} better determines the terminal of answer process compared with \textit{SelfAsk}.}
We analyze the qualitative results for \textit{RGNN-Ret} and \textit{SelfAsk} since they have the similar answering procedures by generating subquestions and use them for retrieval.
The qualitative results of them are shown in Table \ref{tab:case_rgnn}.
For the first question `\textit{Which award the performer of song One More Time (Joe Jackson Song) earned?}', $\textit{SelfAsk}$ generates the first-step subquestion and also answer it correctly.
It obtains the knowledge that `the performer of the song One More Time (Joe Jackson Song) is Joe Jackson'.
However, it terminates the answering process at this step and thus outputs the incorrect final answer.
In contrast, \textit{RGNN-Ret} understands that the generated intermediate answers are not sufficient to output the final answer and thus continues the next-step answering.
Consequently, it can output the final answer `\textit{Grammy}'.
This qualitative comparison between \textit{RGNN-Ret} and \textit{SelfAsk} demonstrates the effectiveness of our proposed \textit{self-critique} technique in determining the termination of the answering process and improving the accuracy for QA.

\textbf{\textit{RGNN-Ret} enhances the retrieval process for subquestions.}
For the second question `\textit{What other notable work did the creator of Shrek make?}' in Table \ref{tab:case_rgnn}, both \textit{SelfAsk} and \textit{RGNN-Ret} are able to correctly answer the first subquestion and generate the second-step subquestion `What other notable works did William Steig create?'.
However, \textit{SelfAsk} cannot retrieve the knowledge about `other work of William Steig', which locates in another passage about the book `Doctor De Soto'.
Using SBERT for retrieval fails to retrieve this passage for QA.
In contrast, \textit{RGNN-Ret} retrieve this passage since it enhances the semantic distance by integrating with the small semantic distances from the previous step.
Consequently, \textit{RGNN-Ret} outputs the correct answer.
This qualitative example demonstrates that the RGNN can enhance the retrieval over steps for subquestions and thus improve the accuracy for QA.

\section{Prompts for Experiments}
\label{prompt}
We display prompt templates of all the methods in this section.

\subsection{Prompt template for {Direct}.}

The prompt template for \textit{Direct} is shown as follows:

\begin{minipage}{0.9\columnwidth}
    \centering
    \begin{tcolorbox}[title= {Prompt template for {Direct} }, label=prompt_direct]
        \small
        Instruction: Given the following question, create a final answer to the question. Please answer less than 6 words.
        \\
        
         \texttt{<Question>}
    \end{tcolorbox}
    \vspace{-1mm}
\end{minipage}
where \texttt{<Question>} indicates the question.

\subsection{Prompt templates for dense retrievers}

The prompt template for retrievers (bm25, DPR, and SBERT) is shown as follows:

\begin{minipage}{0.99\columnwidth}
    \centering
    \begin{tcolorbox}[title= Prompt template for retrievers]
        \small
        Instruction: Given the following question, create a final answer to the question. Please answer less than 6 words.
        \\

        Context:
        
         \texttt{<Context>}
        \\

        Question:
        
         \texttt{<Question>}

    \end{tcolorbox}
    \vspace{-1mm}
\end{minipage}
where \texttt{<Context>} indicates the retrieved passages.

We repeat the question before the length retrieved passages when answering the questions on IIRC and 2WikiMQA datasets for better performance.

\subsection{Prompt templates for IRCoT and KGP}
IRCoT and KGP employ different retrieval methods but the same prompt templates for QA.
The prompt template is shown as follows:

\begin{minipage}{0.99\columnwidth}
    \centering
    \begin{tcolorbox}[title= Prompt template for IRCoT / KGP]
        \small
        Instruction: Given the following question, create a final answer to the question. Please answer less than 6 words.
        \\

        Context:
        
         \texttt{<Context>}
        \\

        Question:
        
         \texttt{<Question>}

    \end{tcolorbox}
    \vspace{1mm}
\end{minipage}
where \texttt{<Context>} indicates the retrieved passages.

We repeat the question before the length retrieved passages when answering the questions on IIRC and 2WikiMQA datasets for better performance.

\subsection{Prompt template for SelfAsk}

SelfAsk prompts LLMs to generate the follow-up question and answers it with retrieved passages until generating the final
answer.
The retrieved passages are included into the prompt when the LLMs are prompted to answer the follow-up question.

\begin{minipage}{0.99\columnwidth}
    \centering
    \begin{tcolorbox}[title= Prompt template for SelfAsk]
        \small
        Instruction: Your goal is to answer the question step by step following procedures of examples. If no follow up questions are necessary, answer the question in five words directly in format: So the final answer is: xxx.
        \\

        Here are the examples of how to answer the questions:
        \\
        \texttt{<Examples>}
        \\

        Context (Optional):
        
         \texttt{<Context>}
        \\
        
        Question:
        \\
         \texttt{<Question>}
    \end{tcolorbox}
    \vspace{1mm}
\end{minipage}

We repeat the question before the length retrieved passages when answering the follow-up questions on IIRC and 2WikiMQA datasets for better performance.

\subsection{Prompt template for ITER-RETGEN}

ITER-RETGEN iteratively answers questions with retrieved passages and uses the generated answer for the next-step retrieval, which continues until the generation of final answer.
The prompt template is shown as follows:

\begin{minipage}{0.99\columnwidth}
    \centering
    \begin{tcolorbox}[title= Prompt template for ITER-RETGEN]
        \small
        Instruction: Given the following question, create a final answer to the question. Please answer less than 6 words.
        \\

        Here are the examples of how to answer the questions:
        \\
        \texttt{<Examples>}
        \\
        
        Context:
        
         \texttt{<Context>}
        \\
        
        Question:
        \\
         \texttt{<Question>}
        \\ 

         Let's think step by step.
    \end{tcolorbox}
    \vspace{1mm}
\end{minipage}

\subsection{Prompt template for RGNN-Ret}
\label{prompt_self_critique}
\textit{RGNN-Ret} iteratively generates the next-step subquestion, answers the subquestion, and performs self-critique until the generation of final answer.
The prompt template of these procedures are shown as follows:

\begin{minipage}{0.99\columnwidth}
    \centering
    \begin{tcolorbox}[title= Prompt template for generating next-step subquestions ]
        \small
        Instruction: Your goal is to ask the next step question logically.
        \\

        Here are the examples of how to answer the questions:
        \\
        \texttt{<Examples>}
        \\
        
        Question:
        \\
         \texttt{<Question>}
         
    \end{tcolorbox}
    \vspace{1mm}
\end{minipage}

\begin{minipage}{0.99\columnwidth}
    \centering
    \begin{tcolorbox}[title= Prompt template for generating subanswers ]
        \small
        Instruction: Your goal is to answer the next step question. I will provide you some wikipedia snippets, and you need to answer the next step question by considering the wikipedia snippets.
        \\

        Context:
        \\
        \texttt{<Context>}
        \\
        
        Question:
        \\
         \texttt{<Question>}
         
    \end{tcolorbox}
    \vspace{1mm}
\end{minipage}

We repeat the question before the length retrieved passages when answering the next-step subanswer on IIRC and 2WikiMQA datasets for better performance.

\begin{minipage}{0.99\columnwidth}
    \centering
    \begin{tcolorbox}[title= Prompt template of self-critique ]
        \small
        Instruction: You are a wikipedia QA expert. Your goal is to critique whether the intermediate answers are enough to generate the final answer to the question.First analyze if the intermediate answers is enough to generate the final answer step by step logically. Then, if it is enough, output 'Critique: yes'. If not, output 'Critique: no'.
        \\
        
        Question:
        \\
         \texttt{<Question>}
        \\
        
         "Analyze if the intermediate answers is enough to generate the final answer to the question step by step logically. Then, if it is enough, output 'Critique: yes'. If not, output 'Critique: no'.
         
    \end{tcolorbox}
    \vspace{1mm}
\end{minipage}

The demonstrations of \textit{RGNN-Ret} for asking the next-step subquestion on MuSiQue, IIRC, and 2WikiMQA datasets are shown as follows:

\begin{minipage}{0.99\columnwidth}
    \centering
    \begin{tcolorbox}[title= Demonstrations of \textit{RGNN-Ret} for asking next-step subquestions on MuSiQue dataset]
        \small
        Question: Who lived longer, Muhammad Ali or Alan Turing?\\
Next step: How old was Muhammad Ali when he died?\\
Intermediate answer: Muhammad Ali was 74 years old when he died.\\
Next step: How old was Alan Turing when he died?\\

Question: When was the founder of craigslist born?\\
Next step: Who was the founder of craigslist?\\
Intermediate answer: Craigslist was founded by Craig Newmark.\\
Next step: When was Craig Newmark born?\\

Question: Who was the maternal grandfather of George Washington?\\
Next step: Who was the mother of George Washington?\\
Intermediate answer: The mother of George Washington was Mary Ball Washington. \\
Next step: Who was the father of Mary Ball Washington?\\

Question: Are both the directors of Jaws and Casino Royale from the same country? \\
Next step: Who is the director of Jaws? \\
Intermediate Answer: The director of Jaws is Steven Spielberg. \\
Next step: Where is Steven Spielberg from? \\
Intermediate Answer: The United States.\\  
Next step: Who is the director of Casino Royale? \\
Intermediate Answer: The director of Casino Royale is Martin Campbell. \\
Next step: Where is Martin Campbell from? 
    \end{tcolorbox}
    \vspace{1mm}
\end{minipage}

\begin{minipage}{0.99\columnwidth}
    \centering
    \begin{tcolorbox}[title= Demonstrations of \textit{RGNN-Ret} for asking next-step subquestions on IIRC dataset]
        \small
        Question: Who lived longer, Muhammad Ali or Alan Turing? \\
Next step: How old was Muhammad Ali when he died?\\
Intermediate answer: Muhammad Ali was 74 years old when he died.\\
Next step: How old was Alan Turing when he died?\\

Question: When was the founder of craigslist born?\\
Next step: Who was the founder of craigslist?\\
Intermediate answer: Craigslist was founded by Craig Newmark.\\
Next step: When was Craig Newmark born?\\

Question: Was the city where Eva was born the capital of its country?\\
Next step: Where was Eva born?\\
Intermediate answer: Eva was born in Berlin, Germany.\\
Next step: Is Berlin the capital of Germany?\\

Question: Was Ryuji Yamakawa a good solo wrestler?\\
Next step: Who is Ryuji Yamakawa?\\
Intermediate answer: Ryuji Yamakawa is a Japanese professional wrestler.\\
Next step: Was Ryuji Yamakawa a good solo wrestler?
    \end{tcolorbox}
    \vspace{1mm}
\end{minipage}

\begin{minipage}{0.99\columnwidth}
    \centering
    \begin{tcolorbox}[title= Demonstrations of \textit{RGNN-Ret} for asking next-step subquestions on 2WikiMQA dataset]
        \small
        Question: Who lived longer, Muhammad Ali or Alan Turing? \\
Next step: How long did Muhammad Ali live? \\
Intermediate answer: Muhammad Ali was 74 years old when he died.\\
Next step: How long did Alan Turing live?\\

Question: Who was the paternal grandfather of Princess Alexandrine Of Prussia (1842-1906)?\\
Next step: Who was the father of Princess Alexandrine Of Prussia (1842-1906)?\\
Intermediate answer: Princess Alexandrine Of Prussia (1842-1906) was the daughter of Prince Albert of Prussia.\\
Next step: Who was the father of Prince Albert of Prussia?\\

Question: Who is the mother of the composer of film 404 (Film)?\\
Next step: Who is the composer of film 404 (Film)?\\
Intermediate answer: The composer of film 404 (Film) is Ilayaraja.\\
Next step: Who is the mother of Ilayaraja?\\

Question: Do both films Across The Badlands and A Gutter Magdalene have the directors that share the same nationality?\\
Next step: What is the nationality of the director of Across The Badlands?\\
Intermediate answer: The director of Across The Badlands is American.\\
Next step: What is the nationality of the director of A Gutter Magdalene?\\
Intermediate answer: The director of A Gutter Magdalene is American.\\
Next step: Do both films Across The Badlands and A Gutter Magdalene have the directors that share the same nationality?\\

Question: What is the home stadium of the team that Asprey hit a hat trick against on 16 January 1961?\\
Next step: Which team did Asprey hit a hat trick against on 16 January 1961?\\
Intermediate answer: Asprey hit a hat trick against Charlton Athletic on 16 January 1961.\\
Next step: What is the home stadium of Charlton Athletic?
    \end{tcolorbox}
    \vspace{1mm}
\end{minipage}

\newpage
\section{Training data for RGNN}
\label{train data}
We manually generate subquestions for 5 questions sampled from the preset training data for MuSiQue, IIRC, and 2WikiMQA datasets.
The training samples are shown as follows:

\begin{minipage}{0.99\columnwidth}
    \centering
    \begin{tcolorbox}[title= Training data of MuSiQue for RGNN]
        \small
        1. Question:
        Why did Roncali leave the place of death of the creator of Malchiostro Annunciation?
        \\
        
        Subquestions:
        
        Who is the creator of Malchiostro Annunciation?

        Where did Titian die?

        Why did Roncali leave Venice?
        \\
        
        2. Question: Where did who argued that the country of citizenship of Victor Denisov had itself beome an imperialist power declare that he would intervene in the Korean conflict?
        \\

        Subquestions:

        What is the country of citizenship of Victor Denisov?

        Who argued that Russia had itself become an imperialist power?

        Where did Mao Zedong declare that he would intervene in the Korean conflict?
        \\
        
        3. Question: What military overran much of Erich Zakowski's place of birth?
        \\
        
        Subquestions:

        What is the place of birth of Erich Zakowski?

        What military overran East Prussia?
        \\
        
        4. Question: How many people were in British Colonies where does the london broil cut come from ?
        \\
        
        Subquestions:

        Where does the london broil cut come from?

        How many people were in North American?
        \\

        5. Question: When was the country established that lies immediately north of the Persian Gulf and the region where the country containing Urim is located?
        \\

        Subquestions:

        What is the region containing Urim?

        Where is the region that Iraq is located?

        What is the country that lies immediately north of the Persian Gulf and the Middle East?

        When was Iran established?
         
    \end{tcolorbox}
    \vspace{1mm}
\end{minipage}

\begin{minipage}{0.99\columnwidth}
    \centering
    \begin{tcolorbox}[title= Training data of IIRC for RGNN]
        \small
        1. Question:
        Was the city where Eva was born the capital of its country?
        \\
        
        Subquestions:
        
        Where was Eva born?

        Is Berlin the capital of its country?
        \\
        
        2. Question: In what state did Galambos attend high school?
        \\

        Subquestions:

        What high school did Galambos attend?

        In what state is Athens High School located?
        \\
        
        3. Question: What was the previous name of the team that Feng started playing with in 1999?
        \\
        
        Subquestions:

        What team did Feng start playing with in 1999?

        What was the previous name of Chongqing Longxin?
        \\
        
        4. Question: How many years after the first Marvel Cinematic Universe film came out was Black Panther released?
        \\
        
        Subquestions:

        When was the first Marvel Cinematic Universe film released?

        When was Black Panther released?

        How many years after Iron Man came out was Black Panther released?
        \\

        5. Question: Who was the first draft pick the year Damarius Bilbo went undrafted?
        \\

        Subquestions:

        What year did Damarius Bilbo go undrafted?

        Who was the first draft pick in 2006?
         
    \end{tcolorbox}
    \vspace{1mm}
\end{minipage}

\begin{minipage}{0.99\columnwidth}
    \centering
    \begin{tcolorbox}[title= Training data of 2WikiMQA for RGNN]
        \small
        1. Question:
        Which film came out earlier, Subliminal Seduction or Australia Marches With Britain?
        \\
        
        Subquestions:
        
        When did Subliminal Seduction come out?

        When did Australia Marches With Britain come out?
        \\
        
        2. Question: Who is the father-in-law of Deuteria?
        \\

        Subquestions:

        Who is the husband of Deuteria?

        Who is the father of Eusebio?
        \\
        
        3. Question: Who is the mother of the composer of film 404 (Film)?
        \\
        
        Subquestions:

        Who is the composer of film 404 (Film)?

        Who is the mother of Ilayaraja?
        \\
        
        4. Question: Which country the composer of film Sergeant Hassan is from?
        \\
        
        Subquestions:

        Who is the composer of film Sergeant Hassan?

        Which country is Tamer Karaoğlu from?
        \\
        
        5. Question: Do both films Across The Badlands and A Gutter Magdalene have the directors that share the same nationality?
        \\

        Subquestions:

        What is the director of Across The Badlands?

        Where is Fred F. Sears from?

        What is the director of A Gutter Magdalene?

        Where is George Melford from?
         
    \end{tcolorbox}
    \vspace{1mm}
\end{minipage}

\end{document}